\documentclass[preprint,12pt]{elsarticle}
\usepackage{algorithm}
\usepackage{algorithmic}

\usepackage{color}
\usepackage{hyperref}

\usepackage[normalem]{ulem}
\useunder{\uline}{\ul}{}

\usepackage{amsmath}
\usepackage{amsthm}
\usepackage{amssymb}
\usepackage{booktabs}

\usepackage{subfig}
\usepackage{float}

\usepackage{amssymb}

\journal{Information Fusion}

\begin{document}
	
	\begin{frontmatter}
		
		
		
		\title{Anchor-based Multi-view Subspace Clustering with Hierarchical Feature Descent}
		
		
		\author{Qiyuan Ou \textsuperscript{a}}
		
		\affiliation{organization={College of Computer, National University of Defense Technology},
			addressline={ouqiyuan14, zhangpei, enzhu@nudt.edu.cn}, 
			city={Changsha},
			postcode={410073}, 
			state={Hunan},
			country={China}}
		
		\author{Siwei Wang \textsuperscript{b}}
		
		\affiliation{organization={Intelligent Game and Decision Lab},
				addressline={wangsiwei13@nudt.edu.cn}, 
				city={Beijing},
				postcode={100000}, 
				country={China}}        
		
		\author{Pei Zhang \textsuperscript{a}}   
		
		\author{Sihang Zhou \textsuperscript{c}} 
		\affiliation{organization={College of Intelligence Science and Technology, National University of Defense Technology},
			addressline={sihangjoe@gmail.com}, 
			city={Changsha},
			postcode={410073}, 
			state={Hunan},
			country={China}}
		
		\author{En Zhu \textsuperscript{a,$\dagger$}}   
		\footnotetext{This work is supported by National Key R\&D Program of China (No. 2022ZD0209103), National Natural Science Foundation of China (No. 62325604, 62276271) and Hunan Provincial Graduate Student Research Program (No.CX20230050).}
		
		\begin{abstract}
			Multi-view clustering has attracted growing attention owing to its capabilities of aggregating information from various sources and its promising horizons in public affairs. Up till now, many advanced approaches have been proposed in recent literature. However, there are several ongoing difficulties to be tackled. One common dilemma occurs while attempting to align the features of different views. {Moreover, due to the fact that many existing multi-view clustering algorithms stem from spectral clustering, this results to cubic time complexity w.r.t. the number of dataset. However, we propose Anchor-based Multi-view Subspace Clustering with Hierarchical Feature Descent(MVSC-HFD) to tackle the discrepancy among views through hierarchical feature descent and project to a common subspace( STAGE 1), which reveals dependency of different views. We further reduce the computational complexity to linear time cost through a unified sampling strategy in the common subspace( STAGE 2), followed by anchor-based subspace clustering to learn the bipartite graph collectively( STAGE 3). }
			Extensive experimental results on public benchmark datasets demonstrate that our proposed model consistently outperforms the state-of-the-art techniques. {Our code is publicly available at \url{https://github.com/QiyuanOu/MVSC-HFD/tree/main}.}
			
		\end{abstract}
		
		
		\begin{highlights}
			\item MVSC-HFD has high adaptability to {various dimensions of different views}.
			\item {Tackle the discrepency among different views.}
			\item Reduce multiple views of different dimensions hierarchically to {a common subspace}.
			\item Uniformly learn anchors in common subspace and construct bipartite graphs.
			\item Low spatial and time consumption which are linear to data size. 
		\end{highlights}
		
		\begin{keyword}
			Multi-view Clustering\sep multimodal fusion\sep subspace clustering\sep anchor graph\sep Machine Learning
			
			
			
			
		\end{keyword}
		
	\end{frontmatter}
	
	

	\section{Introduction}

	With the advance of information technology, many practical datasets are collected from different channels or provided in different modalities\cite{cai2023seeking,liang2022robust,yang2022fast,DBLP:journals/corr/abs-1304-5634}. As a result, multi-view learning has gained increasing attention due to its ability in handling and learning from various source information. Among those learning datasets, the majorities are unlabeled with only features of different views attached to them. {Many clustering algorithms have been used in practice in order to discover inherent patterns from these cluttered datasets and also to reveal things that were previously unknown to humans.} {One common task that multi-view clustering algorithms confront is that in some datasets, different views vary tremendously in dimensions, which will adversely affect the extraction and expression of common representations and data patterns revealed by features from each view\cite{DBLP:journals/inffus/YanGR0LXL23,DBLP:conf/cvpr/YanZLT0LL23,DBLP:journals/nn/TangSTZLHZ23,DBLP:conf/mm/YanXLYT22,DBLP:journals/chinaf/TangZZLZZ23,DBLP:journals/tkde/TangZLZZXW22,DBLP:journals/tkde/TangLWLZZ23}.}
	
	Projecting different views into a common space is a long-term academic issue, e.g., Canonical Correlation Analysis (CCA)\cite{hotelling1992relations} and its variants \cite{DBLP:journals/corr/abs-cs-0609071,DBLP:conf/icml/AndrewABL13} aim to align a pair of views with linear correlation through complex nonlinear transformation. This technique is restricted to the simple case where only two views are provided. In multi-view circumstances, \cite{wang2020robust} assumes that the high dimensional data emanate from noisy redundancies, which can be eliminated by minimizing $l_{2,1}$ norm of the representation. It is a typical phenomenon that, the lower the dimensions of data points are reduced to, the more the information is lost \cite{DBLP:journals/bioinformatics/QaqishOHC17,DBLP:journals/tkde/WeiYFCL23}. On the contrary, the higher dimensions we preserve to learn from, the better the resulting representation captures the underlying semantic essence of the data points, although with typically higher computational cost. 
	
	Subspace clustering directly bypasses the curse of dimensionality\cite{DBLP:journals/ior/Bellman54} through expressing each data point as a linear combination of other points and minimizing the reconstruction loss to attain the combination coefficient\cite{DBLP:journals/pami/LiuLYSYM13,DBLP:journals/pami/ElhamifarV13}. Recently, subspace clustering has been extended to multi-view circumstances\cite{DBLP:conf/iccv/GaoNLH15,DBLP:conf/cvpr/CaoZFLZ15,yan2023collaborative,chen2022diversity,zhao2023auto}, where each view owns a distinct coefficient matrix and a regularization term is used to obtain the joint coefficient matrix, which is then expressed as similarity matrix for downstream spectral clustering and k-means clustering tasks. It is obvious that, as the self-representation matrix has the size of $n \times n$, which is the same as the scale of data points, extremely high computational and spatial cost are imposed on the optimization procedure and the clustering task, which would bring $\mathcal{O}(n^3)$ time cost and at least $\mathcal{O}(n^2)$ storage cost, respectively. As a typical paradigm for reducing the computational and spatial cost, the anchor scheme is introduced to various clustering models\cite{chen2011large,wang2016scalable,kang2020large,zhou2021multiple,shu2022self,yang2023robust}. It is a common practice to carefully choose (mainly adopt k-means clustering as a preliminary scheme) or randomly sample from the original dataset. The above formula bears several drawbacks: a) Results highly rely on the initialization as the anchors are predetermined. Although a preliminary $k$-means clustering on data features can make a favorable selection, {the pre-selected anchor points are fixed in the same dimensional space as the original data, so the practicability can be greatly reduced when the nodes are not linearly separable}; b) In multi-view learning background, {the fundamental problem with most of the frameworks is that both the anchors and the graph reconstruction matrices are view-specific, which ignores the alliance between different views;} c) For final spectral clustering step, the fusion step is required to build the global graph based on view-specific graphs. In this circumstance, the fusion method will also extensively affect the clustering result. {d) Specifically for \cite{wang2021fast,li2020multi}, which efficiently learn anchors using subspace clustering method, assume that cross-view consistent bipartite graph exists. Accordingly, they deploy dependency among views without tackling the discrepancy of dimensions and features adequately, thus inhibiting a better performance.}
	
	To address the aforementioned limitations, we propose a novel approach named Anchor-based \textbf{M}ulti-\textbf{v}iew \textbf{S}ubspace \textbf{C}lustering with \textbf{H}ierarchical \textbf{F}eature \textbf{D}escent(MVSC-HFD). Specifically, in our method, we adopt three mechanisms to extract the bipartite graph reconstruction matrix. First, the input data points of different views are reduced to the same dimensional space through hierarchical dimensional projection matrices. Second, we assume that for all views, data points sketch the same structure in this unified dimension, and can be jointly expressed by common subspace, thus reuniting the separate clustering of distinct views into a collectively learned framework. Thirdly, to negotiate between feature preservation and efficiency, we adopt an anchor graph with carefully designed dimensions and the number of anchor points. 
	
	Compared with existing methods, our algorithm has the following advantages:{
		\begin{itemize}
			\item[1.] \textbf{High adaptability to various dimensions.} Our algorithm can cater to different dimensions of multiple views.
			\item[2.] \textbf{Reduce the discrepancy and reveal dependency of different views.} Our method can tackle discrepancy among multiple views by hierarchically dimensional descent at different rates into a common subspace for clustering, which increases the effectiveness of the anchors-based clustering scheme.
			\item[3.] \textbf{Model integrity and performance gains.} Aforementioned variables including weight vector, projection matrix, anchor matrix and bipartite graph are collectively learned throughout the optimization procedure with guaranteed convergence. The extensive experiments show the validity of our proposed method from both accuracy and efficiency. 
		\end{itemize}
	}


	\section{Related Work}
	In this section, we precede our work with some most relevant work, including non-negative matrix factorization, subspace clustering and multi-view subspace clustering. For clarity, to elaborate on the correlation and inspiration of the preceding work, we use the unified annotations throughout this prologue, which is shown below.
	
	\begin{table}[H]\small
		\centering
		\caption{Summary of notations}\label{notations}
		\begin{tabular}{c|c}
			\toprule
			$k$ & The number of data clusters. \tabularnewline
			$p$ & The number of views. \tabularnewline
			$m$ & The number of anchor points. \tabularnewline
			$\delta$ & The depth of hierarchical feature descent. \tabularnewline
			$d_v$ & Dimension of the $v$-th view.\tabularnewline
			$l_i$ & Dimension of the $i$-th layer.\tabularnewline
			$\boldsymbol{\alpha}\in \mathbb{R}^{p}$ &A concatenation of $p$ weights. \tabularnewline
			$\mathbf{X}^{(v)} \in \mathbb{R}^{d_v \times n}$ & The data matrix of the $v$-th view. \tabularnewline
			$\mathbf{I}_n \in \mathbb{R}^{n \times n}$ & The $n$-th order identity matrix. \tabularnewline
			$\mathbf{W}^{(v)}_{i}$ & The hierarchical anchor graph of the $i$-th depth. \tabularnewline
			$\mathbf{H} \in \mathbb{R}^{k \times n}$ & The partition matrix of consensus embedding. \tabularnewline
			$\mathbf{A} \in \mathbb{R}^{k \times m}$ & The unified anchor point representation. \tabularnewline
			$\mathbf{Z} \in \mathbb{R}^{m \times n}$ & The consensus reconstruction matrix.\tabularnewline
			\bottomrule
		\end{tabular}
	\end{table}
	
	\subsection{Non-Negative Matrix Factorization}
	
\begin{figure}[t]
		\centering
		\includegraphics[width=7.6cm]{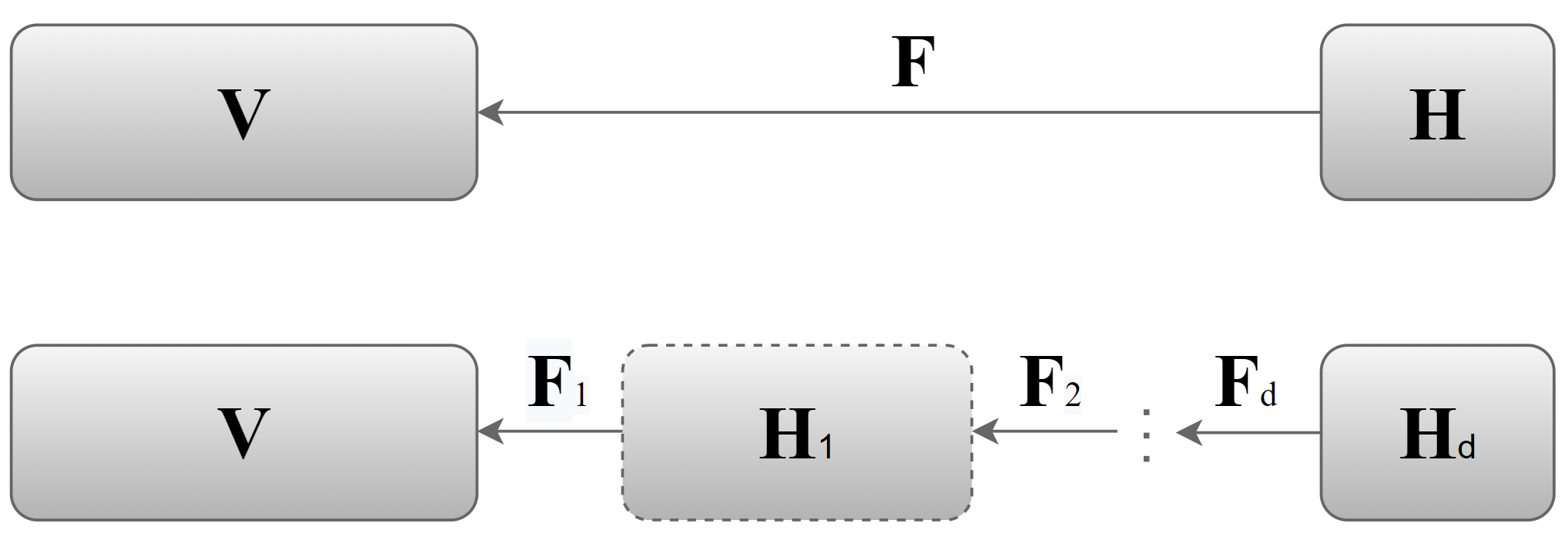}
		\caption{The conventional NMF model applies a linear transformation to the original input space, which is usually of higher dimension. On the other hand, Deep NMF goes a step further by learning multiple layers of hidden representations that gradually reveal the ultimate lower-dimensional representation of the data in an hierarchical manner.}\label{fig:deep}
	\end{figure}

	Non-negative matrix factorization (NMF or NNMF) is a group of algorithms that factorize a matrix $\mathbf{V}$ into (usually) two matrices $\mathbf{F}$ and $\mathbf{H}$, with the uniform constraint that all these three matrices have the property of non-negativity, which makes the formulation easy to solve whereas endow the practicality of the algorithms in real-world applications such as processing of natural language, audio-visual signal spectrograms or missing data imputation. 
	\begin{align}
		\mathbf{V} \approx \mathbf{FH},
	\end{align}
	{where $\mathbf{V} \in \mathbb{R}^{d \times n}$ is the data matrix, each column represents a vector of dimension $d$ and there are $n$ numbers of such vector. $\mathbf{F} \in \mathbb{R}^{d \times k}$ and $k$ is the number of base vectors. $\mathbf{H} \in \mathbb{R}^{k \times n}$ is the representation mapped through the mapping matrix $\mathbf{F}$. While keeping the information unchanged as much as possible, NMF aims to simplify high dimensional data $\mathbf{V}$ to low dimensional mode $\mathbf{H}$, in order to estimate the essential structure which $\mathbf{V}$ represents. Original NMF and Deep-NMF are illustrated in Fig.~\ref{fig:deep}.}
	
	\subsection{Subspace Clustering}
	Subspace clustering assumes that the whole dataset can be represented linearly by a fraction of them, which is called subspace. By minimizing the reconstruction loss, the target function acquires a combination coefficient matrix. Considerable advancements have been achieved in uncovering underlying subspaces. Currently, there are two main research directions in subspace clustering: sparse subspace clustering (SSC) \cite{DBLP:journals/pami/ElhamifarV13} and low-rank representation (LRR)\cite{DBLP:journals/pami/LiuLYSYM13}. SSC employs sparse representation techniques to cluster a set of multi-subspaces, while LRR utilizes the nuclear norm to impose an approximate low-rank constraint on the reconstruction matrix. Both types of algorithms follow a similar routine: they first reconstruct the data using the original input and then apply spectral clustering on the resulting similarity graph. For a given data matrix $\mathbf{X} \in \mathbb{R}^{d \times n}$ with $n$ samples and $d$ features, the formula for naive single-view subspace clustering is presented below, 
	\begin{align}
		\nonumber
		& \min _{\mathbf{S}}\|\mathbf{X}-\mathbf{X} \mathbf{S}\|_{\mathrm{F}}^{2}+\lambda \|\mathbf{S}\|_{\dagger} \\
		& s.t.  ~\mathbf{S} \geq \mathbf{0}, ~\mathbf{S}^{\top} \mathbf{1}=\mathbf{1}, ~diag(\mathbf{S})=\boldsymbol{0},
		\label{Algorithm_SC}
	\end{align}
	where $\mathbf{S} \geq \mathbf{0}$ means each element in similarity graph matrix $\mathbf{S}$ is non-negative, and $diag(\mathbf{S})=\boldsymbol{0}$ restricts the diagonal elements to zero, so as to prohibits trivial solution. The term $\|\cdot\|_{\dagger}$ represents a manifold of adequate norms that can serve as regularizers to adjust to various algorithmic application backgrounds. For example, $\ell_1$-norm, $\ell_{21}$-norm, nuclear norm, Frobenius norm are corresponding to different properties like sparsity of matrix, sparsity of row space, low-rank matrices, respectively. $\lambda$ is a regularization parameter to negotiate between the regularization term $\|\cdot\|_{\dagger}$ and the alignment term( the squared Frobenius norm of reconstruction). The constraint $\mathbf{S}^\top \mathbf{1}=\mathbf{1}$ is synonymous with $\sum_{i} \mathbf{S}_{i, :}=1$, ensuring that every data point is assigned to at least one subspace. In order to unveil the inherent correlations among data samples, these algorithms incorporate $\mathbf{S}$ to linearly reconstruct the data points, aligning them with their original counterparts. Hence, $\mathbf{S}$ is often referred to as a reconstruction coefficient matrix or self-representation matrix. A multitude of relative researches build the similarity matrix $\hat{\mathbf{S}}$ as
	\begin{align}
		\hat{\mathbf{S}} = \frac{\mathbf{S}+\mathbf{S}^{\top}}{2},
	\end{align}
	where $\hat{\mathbf{S}}$ is the input non-negative symmetric similarity matrix of the downstream tasks, usually with spectral clustering methods which generate the result of data cluster as follows, 
	
	\begin{align}
		&\max _{\mathbf{H}} \operatorname{Tr}\left(\mathbf{H}^{\top} \hat{\mathbf{S}} \mathbf{H}\right), 
		&s.t. ~\mathbf{H} \in \mathbb{R}^{n \times k}, ~\mathbf{H}^{\top} \mathbf{H}=\mathbf{I}_{k},
	\end{align}
	
	where $\operatorname{Tr}(\cdot)$ means the trace of the matrix. $\mathbf{H}$ is known as the spectral embedding of data matrices, which is then used as the input of the k-means clustering scheme to get the final cluster indicating matrix. 
	

	\subsection{Multi-view Subspace Clustering}
	In multi-view circumstances, suppose $\mathbf{X}$ represents multi-view data matrix with $\left\{\mathbf{X}_{i}\right\}_{i=1}^{p}=\left[\mathbf{x}_{i}^{(1)}, \mathbf{x}_{i}^{(2)}, \ldots, \mathbf{x}_{i}^{(n)}\right] \in \mathbb{R}^{d_{i} \times n}$, where $d_i$ means the feature dimension of the $i$-th view and $n$ means sample number. Multi-view subspace clustering scheme can be represented below:
	\begin{align}
		\nonumber
		& \min _{\mathbf{S}_i,\mathbf{S}}\sum_{i=1}^{p}\|\mathbf{X}_i-\mathbf{X}_i \mathbf{S}_i\|_{\mathrm{F}}^{2}+\lambda \|\mathbf{S}_i\|_{\dagger}+\Omega(\mathbf{S},\mathbf{S}_i) \\
		& s.t.  ~\mathbf{S,S_i} \geq \mathbf{0}, ~\mathbf{S}^{\top} \mathbf{1}=\mathbf{1}, ~\mathbf{S}_i^{\top} \mathbf{1}=\mathbf{1}, ~diag(\mathbf{S})=\boldsymbol{0}, ~diag(\mathbf{S}_i)=\boldsymbol{0},
		\label{Algorithm_MVSC}
	\end{align}
	where the regularization term $\|\cdot\|_{\dagger}$ endows view-specific anchor graphs $\{\mathbf{S}_i\}_{i=1}^{p}$ with different properties. $\mathbf{S}$ is the self-representation matrix optimized by minimizing the term $\Omega(\mathbf{S},\mathbf{S}_i)$, which represents categories of many fusion techniques. The clustering process of existing methods for multi-view subspace clustering is in two stages:
	- Step 1: Learning the subspace representation and integrating multi-view information at the level of similarity matrix or self-expression matrix by fusing the graphs or self-expression matrices of multiple views (reconstruction matrix) to form a shared similarity matrix \cite{DBLP:conf/aaai/LiNHH15,DBLP:journals/tip/WangLWZZH15,DBLP:journals/tnn/WangWLG18}. For example, \cite{DBLP:conf/aaai/Guo13} learned shared sparsity by performing Matrix Factorization (MF) of the subspace representations. Unlike obtaining shared self-expressions or shared graphs directly, to better exploit the complementary information and diversity between different views, the work \cite{DBLP:conf/cvpr/CaoZFLZ15,DBLP:conf/mm/HuangTX0L21} and \cite{DBLP:conf/aaai/XiaPDY14} were performed by Hilbert-Schmidt independence criterion (HSIC) and Markov chain, respectively. Wang et al.\cite{DBLP:conf/cvpr/WangGLZL17} introduced an exclusion constraint to learn different subspace representations in each view, which force multiple subspace representations to be sufficiently different from each other, and then added them together directly or adaptively to obtain a uniform graph or self-expression matrix.
	- In the second step: Clustering, the obtained uniform graph representation is fed into an off-the-shelf clustering model (generally using spectral clustering) to obtain the final clustering results.
	
	Although satisfying clustering performances have been achieved by existing multi-view subspace learning algorithms in many real world applications, the large computational consumption and storage cost by optimizing the $n\times n$ reconstruction matrix $\mathbf{S}$ prohibits the scalability of these methods on large-scale background.
	
	%

	\section{Methodology}
	
	In the section that follows, we first sketch our motivation and the outline of our proposed method(MVSC-HFD). Then we elaborate on the optimization of the algorithm with proved convergence. Finally, we summarize the algorithmic procedure and analyze the computatioanl complexity of the learning task. 
	
	\subsection{Motivation}
\begin{figure}
		\centering
		\includegraphics[width=0.99\textwidth]{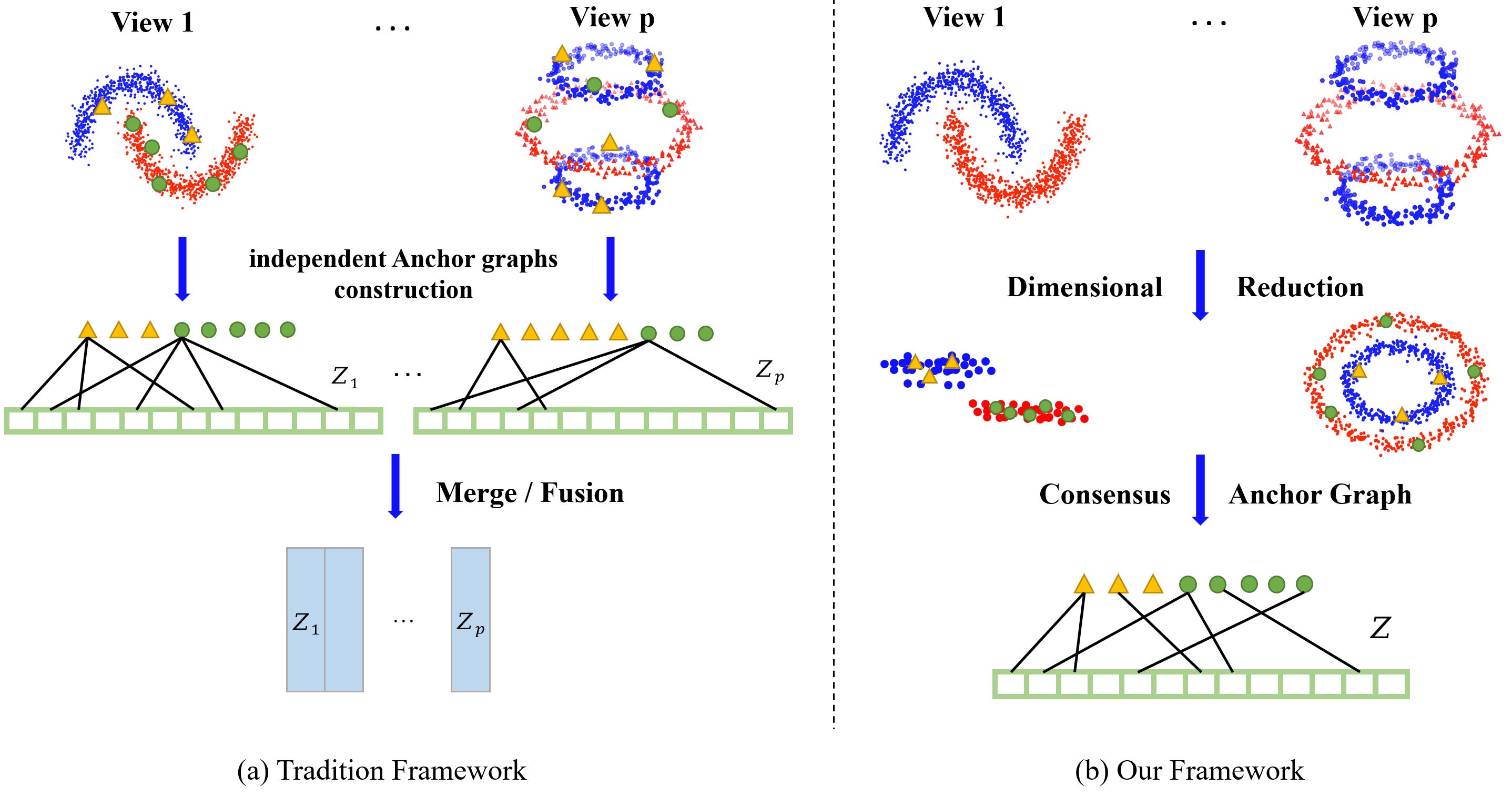}
		\caption{The frameworks of the traditional methods and ours.}
		\label{fig:frame}
	\end{figure}

	In large-scale data circumstances (either with large data size or with high dimension), the time and space overhead of traditional learning algorithms would be overwhelmingly huge\cite{DBLP:phd/ndltd/Shakhnarovich05}. For high dimensional datasets, it is of much redundancy to directly learn representations from original data samples. Although subspace clustering provides an alternative approach to circumvent high-dimensional feature learning, it still spends high computational and storage costs to construct global similarity graphs between each two elements. And it is obvious that a large proportion of the data does not contribute significantly to the clustering structure. Anchor graph theory\cite{wang2016scalable,yang2022multiview} proposes that a good clustering structure can be acquired with only a fraction of the sample points by correlating these sample points with the full sets of data. These similarity matrices are also termed as anchor graph or bipartite graph. Nevertheless, existing anchor-based subspace learning methods often use delicately chosen landmarks beforehand, hence the clustering result highly relies upon these pre-specified knowledge, which restricts the scalability of the algorithm. {We propose an adaptable clustering algorithm to tackle not only high dimensional data, but also large-scale samples by hierarchically projecting the original data to a low dimensional common subspace, where a set of anchors are learned uniformly to construct the bipartite graph.} We demonstrate the overall procedure together with traditional methods in fig.~\ref{fig:frame}.

	\subsection{The Proposed Formulation}
	
	\begin{center}
		\begin{equation}\label{target1}
			\begin{split}
				&\mathop{\min}\limits_{\small \boldsymbol{\alpha},\{\mathbf{W}^{(v)}_{i}\}_{i=1}^{\delta},\mathbf{A},\mathbf{Z}} \large \sum_{v=1}^{p} \alpha_{v}^{2} \| \mathbf{X}^{(v)}-\mathbf{W}^{(v)}_{1}\mathbf{W}^{(v)}_{2}\cdots\mathbf{W}^{(v)}_{\delta}\mathbf{A}\mathbf{Z}\| _{F}^2, \\
				&s.t.~ \small\boldsymbol{\alpha}^{\top} \mathbf{1}=1,
				{\mathbf{W}^{(v)}_{i}}^{\top}\mathbf{W}^{(v)}_{i} = \mathbf{I}_{l_{i}}, 
				\mathbf{A}^{\top}\mathbf{A} = \mathbf{I}_m, \\
				&~~~~~~\mathbf{Z} \geq 0, \mathbf{Z}^{\top}\mathbf{1} = \mathbf{1},
			\end{split}
		\end{equation}
	\end{center}
	
	where $\mathbf{X}^{(v)} \in \mathbb{R}^{d_{v} \times n}$ is the original data matrix of the $v$-th view, with $d_v$ denoting the dimension and $n$ denoting the number of data points. The hierarchical dimensional projection matrix $\mathbf{W}^{(v)}_i \in \mathbb{R}^{l_{i-1} \times l_{i}}$ ($l_{0} = d_{v}$, $l_{\delta} = k$, in particular) is the projection matrix that performs dimensional reduction. {Specifically, data in different views often have different dimensions. So usually $l_i$ are not the same for $\mathbf{W}^{(v)}_i$ of different views, which means different dimensional reduction process. Distinctive $\mathbf{W}^{(v)}_i$ represents the discrepancy of different views and the serial multiplier $\mathbf{W}^{(v)}_1\mathbf{W}^{(v)}_2\cdots\mathbf{W}^{(v)}_{\delta}$ aims to reduce the discrepancy.} $\mathbf{A} \in \mathbb{R}^{k \times m}$ is the unified anchor points in which $k$ represents the dimension and $m$ stands for the number of the anchor points. {This anchor matrix $\mathbf{A}$ is shared among different views and reveals the dependency. It represents the common features extracted through dimensional projection matrix multiplier $\mathbf{W}^{(v)}_1\mathbf{W}^{(v)}_2\cdots\mathbf{W}^{(v)}_{\delta}$.} The learnable vector of parameters $\boldsymbol{\alpha}=[\alpha_1,\alpha_2,\cdots,\alpha_p]^{\top}$ represents the weights adherent to different views. And the linear constraint of $\boldsymbol{\alpha}$ corresponds to the quadratic optimization target, which is a convex optimization problem owing to the inherently non-negative nature of Frobenius norm. $\mathbf{Z} \in \mathbb{R}^{m \times n}$ represents the consensus bipartite graph amongst different views, which constructs the pairwise relationship between original data samples and delicately learned landmarks. Traditional subspace clustering algorithms utilize the learned bipartite graph to construct the full similarity graph $\mathbf{S}$ using $\mathbf{S} = \mathbf{\hat{Z}^{\top} \hat{Z}}$ where $\mathbf{\hat{Z}}=\mathbf{Z} \mathbf{\Sigma}^{-1 / 2}$ and $\mathbf{\Sigma}$ denotes a diagonal matrix with $\mathbf{\Sigma}_{j j}=$ $\sum_{i=1}^n \mathbf{Z}_{i j}$. Afterwards, they perform spectral clustering with the acquired similarity matrix to obtain final clustering results. These give rise to $\mathcal{O}(n^3)$ computational consumption and $\mathcal{O}(n^2)$ spatial cost. We adopt a different approach by proving an equivalence in right singular matrix of bipartite graph $\mathbf{Z}$ and eigenvectors of the similarity matrix $\mathbf{S}$: 
	\begin{center}
		\begin{equation}
			\begin{split}
				\mathbf{S} = \mathbf{\hat{Z}^{\top} \hat{Z}}&= (\left(\mathbf{U}_\mathbf{Z} \mathbf{\Lambda}_\mathbf{Z} \mathbf{V}_\mathbf{Z}^{\top}\right)\mathbf{\Sigma}^{-1 / 2})^{\top}\left(\mathbf{U}_\mathbf{Z} \mathbf{\Lambda}_\mathbf{Z} \mathbf{V}_\mathbf{Z}^{\top}\mathbf{\Sigma}^{-1 / 2}\right) \\
				& =\mathbf{V}_\mathbf{Z} \mathbf{\Lambda}_\mathbf{Z}\mathbf{\Sigma}^{-1 / 2}\left(\mathbf{U}_\mathbf{Z}^{\top} \mathbf{U}_\mathbf{Z}\right) \mathbf{\Sigma}^{-1 / 2}\mathbf{\Lambda}_\mathbf{Z} \mathbf{V}_\mathbf{Z}^{\top} \\
				& =\mathbf{V}_\mathbf{Z} \mathbf{\Lambda}_\mathbf{Z}^2 \mathbf{\Sigma}^{-1}\mathbf{V}_\mathbf{Z}^{\top},
			\end{split}
		\end{equation}
	\end{center}
	{where $\mathbf{Z} = \left(\mathbf{U}_\mathbf{Z} \mathbf{\Lambda}_\mathbf{Z} \mathbf{V}_\mathbf{Z}^{\top}\right)$ denotes the singular value decomposition(SVD) of the consensus bipartite graph $\mathbf{Z}$.} Therefore, we only need to compute the right singular matrix of $\mathbf{Z} \in \mathbb{R}^{m \times n}$ for further clustering task (lite $k$-means for example), and this costs $\mathcal{O}(nm^2)$.

	\subsubsection{Optimization Algorithm}
	The decision variables for the objective function are $\{\mathbf{W}^{(v)}_i\}_{i=1}^{\delta},\mathbf{A}$, $\mathbf{Z}$ and $\boldsymbol{\alpha}$. As a result, we introduce a four-step iterative optimization method to approximate the optimal target mentioned in Eq.~\eqref{target1}. In this approach, we keep three variables fixed while optimizing the remaining ones. The specific procedure is derived from the following detailed steps.
	
	\noindent{\textbf{Initialization}.}
	Our algorithm adopts a straightforward initialization by setting $\mathbf{Z}$ as an $m \times n$ matrix with the first $m \times m$ block deposited as identity matrix and the rest elements deposited to be $0$. The anchor matrix $\mathbf{A}$ is initialized as identity matrix w.r.t. its dimension. And $\boldsymbol{\alpha}$ is initialized using mean values factorized by view count. The other decision variable $\{\mathbf{W}^{(v)}_i\}_{i=1}^{\delta}$ is neglected for initializing as it is the first to optimize.
	
	\noindent{\textbf{Update $\mathbf{W}^{(v)}_o$ with $\{\mathbf{W}^{(v)}_i\}_{i=1, i\neq o}^{\delta}$, $\mathbf{A}$, $\mathbf{Z}$ and $\boldsymbol{\alpha}$ fixed.}} Given $\mathbf{A}$, $\mathbf{Z}$, $\boldsymbol{\alpha}$ and the rest $\delta-1$ numbers of $\{\mathbf{W}^{(v)}_i\}_{i=1, i\neq o}^{\delta}$, the optimization problem in Eq. \eqref{target1} with respect to $\mathbf{W}^{(v)}_o$ becomes: 
	
	\begin{align}\label{op_W1}
		\nonumber
		&\min _{\mathbf{W}^{(v)}_{o}} \sum_{v=1}^{p} \alpha_{v}^{2}\left\|\mathbf{X}^{(v)}-\boldsymbol{\Omega}\mathbf{W}^{(v)}_{o}\hat{\mathbf{A}}_{o}\mathbf{Z}\right\|_{F}^{2}, \\
		& s.t. ~{\mathbf{W}^{(v)}_o}^{\top}\mathbf{W}^{(v)}_o = \mathbf{I}_{l_{o}}, 
	\end{align}
	
	Where $\boldsymbol{\Omega}$ denotes the hierarchical dimensional projection series $\mathbf{W}^{(v)}_1\mathbf{W}^{(v)}_2\cdots\mathbf{W}^{(v)}_{o-1}$. The multiplication term $\hat{\mathbf{A}}_{o}\mathbf{Z}$ refers to the anchor-based subspace reconstruction of $i$-th layer's representation, and $\hat{\mathbf{A}}_{o}$ denotes the generalized anchors. Through deduction of matrix Frobenius norm and let $\mathbf{M}^{(v)}=\boldsymbol{\Omega}^{\top}\mathbf{X}^{(v)}\mathbf{Z}^{\top}\mathbf{A}^{\top}$, Eq. \eqref{op_W1} can be reformulated as: 
	\begin{equation}\label{op_W2}
		\max _{\mathbf{W}^{(v)}_{o}} \operatorname{Tr}\left(\mathbf{M}^{(v)}{\mathbf{W}^{(v)}_{o}}^{\top} \right),
		s.t. ~{\mathbf{W}^{(v)}_o}^{\top}\mathbf{W}^{(v)}_o = \mathbf{I}_{l_{o}},
	\end{equation}
	
	{Denote the SVD of matrix $\mathbf{M}^{(v)}$ as $\mathbf{U}_\mathbf{M}\mathbf{D}_\mathbf{M}\mathbf{V}_\mathbf{M}^{\top}$ and the optimal solution of Eq. \eqref{op_W2} is demonstrated as ${\mathbf{W}^{(v)}_o}^{\star}$. In this case, $\mathbf{D}_\mathbf{M}$ is a nonnegative diagonal matrix with each entry representing the singular value of $\mathbf{M}^{(v)}$, both $\mathbf{U}_\mathbf{M}$ and $\mathbf{V}_\mathbf{M}$ are unitary matrix with orthogonal entries. According to deduction in \cite{ou2020anchor}, a closed-form resolution can be found by calculating ${\mathbf{W}^{(v)}_o}^{\star}=\mathbf{U}_\mathbf{M}sign(\mathbf{D}_\mathbf{M})\mathbf{V}_\mathbf{M}^{\top}$, where $sign(\cdot)$ is a function imposed on matrix with each element, which returns $1$ if the element is larger than $0$. And it is obvious that in our case, $sign(\mathbf{D}_\mathbf{M})$ is an identity matrix. Accordingly, the optimal solution can be calculated as:
		\begin{equation}\label{op_W_end}
			{\mathbf{W}^{(v)}_o}^{\star}=\mathbf{U}_\mathbf{M}\mathbf{V}_\mathbf{M}^{\top}
		\end{equation}
	}
	By following the updating expression in Eq. \eqref{op_W_end}, our algorithm updates $\mathbf{W}^{(v)}_i,~i=1,2,\cdots,\delta$ in succession.
	
	\noindent{\textbf{Update $\mathbf{A}$ with $\{\mathbf{W}^{(v)}_i\}_{i=1}^{\delta}$, $\mathbf{Z}$ and $\boldsymbol{\alpha}$ fixed.}}
	When $\boldsymbol{\alpha}$ and $\{\mathbf{W}^{(v)}_i\}_{i=1}^{\delta}$ of each view are all prescribed, updating $\mathbf{A}$ with fixed $\mathbf{Z}$, which relocates the unified anchor matrix, can be termed as: 
	
	\begin{center}
		\begin{equation}\label{op_A1}
			\begin{split}
				&\mathop{\min}\limits_{\small \mathbf{A}} \large \sum_{v=1}^{p} \alpha_{v}^{2} \| \mathbf{X}^{(v)}-\mathbf{W}^{(v)}_{1}\mathbf{W}^{(v)}_{2}\cdots\mathbf{W}^{(v)}_{\delta}\mathbf{A}\mathbf{Z}\| _{F}^2, \\
				&s.t.~ \mathbf{A}^{\top}\mathbf{A} = \mathbf{I}_m,
			\end{split}
		\end{equation}
	\end{center}
	
	Likewise, the deduction for the optimization of the anchor selection variable $\mathbf{A}$ is similar to Eq. \eqref{op_W2}, which can be equally written as:
	\begin{equation}\label{op_A2}
		\max _{\mathbf{A}} \operatorname{Tr}\left(\Phi \mathbf{A}^{\top} \right) s.t. \mathbf{A}^{\top} \mathbf{A}=\mathbf{I}_{m},
	\end{equation}
	where $\Phi=\sum_{v=1}^{p} \alpha_{v}^{2} {\mathbf{W}^{(v)}_{\delta}}^{\top} {\mathbf{W}^{(v)}_{\delta-1}}^{\top} \cdots {\mathbf{W}^{(v)}_{1}}^{\top} \mathbf{X}^{(v)} \mathbf{Z}^{\top}$ and $\Phi \in \mathbb{R}^{k \times m}$.
	
	Corresponding to the updating rule in \cite{ou2020anchor}, it is straightforward to obtain the optimal solution for $\mathbf{A}$ as in Eq. \eqref{op_W2}. The multiplication of the left and right singular matrices of $\Phi$ costs $\mathcal{O}(km^2)$.
	
	\noindent{\textbf{Update $\mathbf{Z}$ with $\mathbf{A}$, $\{\mathbf{W}^{(v)}_i\}_{i=1}^{\delta}$ and $\boldsymbol{\alpha}$ fixed.}}
	Given $\{\mathbf{W}^{(v)}_i\}_{i=1}^{\delta}$, the anchors are separately lodged in $p$ different anchor spaces of the same dimension $k$ and are reunited by the optimized anchor matrix $\mathbf{A}$, which is shared amongst views. This subtask of updating $\mathbf{Z}$ is equivalent to the process of constructing the bipartite graph from the anchor space to the original feature space. The optimization problem in Eq. \eqref{target1} concerning $\mathbf{Z}$ can be written as:
	
	\begin{center}
		\begin{equation}\label{op_Z1}
			\begin{split}
				&\mathop{\min}\limits_{\small \mathbf{Z}} \large \sum_{v=1}^{p} \alpha_{v}^{2} \| \mathbf{X}^{(v)}-\mathbf{W}^{(v)}_{1}\mathbf{W}^{(v)}_{2}\cdots\mathbf{W}^{(v)}_{\delta}\mathbf{A}\mathbf{Z}\| _{F}^2, \\
				&s.t.~ \mathbf{Z} \geq 0, \mathbf{Z}^{\top}\mathbf{1} = \mathbf{1},
			\end{split}
		\end{equation}
	\end{center}
	
	According to the definition of Frobenius norm, we can easily derive Eq. \eqref{op_Z1} as follows:
	
	\begin{center}
		\begin{equation}\label{op_Z2}
			\begin{split}
				&\mathop{\min}\limits_{\small \mathbf{Z}} \sum_{v=1}^{p} \alpha_{v}^{2} \mathrm{Tr} (\mathbf{Z}^{\top}\mathbf{Z}-2{\mathbf{X}^{(v)}}^{\top}\mathbf{P}\mathbf{Z}+{\mathbf{X}^{(v)}}^{\top}\mathbf{X}^{(v)}) \\
				&s.t.~ \mathbf{Z} \in \mathbb{R}^{m \times n}, \mathbf{Z} \geq 0, \mathbf{Z}^{\top} \mathbf{1}=\mathbf{1}, 
			\end{split}
		\end{equation}
	\end{center}
	
	where $\mathbf{P}=\mathbf{W}^{(v)}_{1}\mathbf{W}^{(v)}_{2}\cdots\mathbf{W}^{(v)}_{\delta}\mathbf{A}$. As the optimisation of the columns in $\mathbf{Z}$ is identical with each other, the optimisation task can be separated into $n$ sub-tasks by the identical form as:
	
	\begin{center}
		\begin{equation}\label{op_Z3}
			\begin{split}
				&\min\limits_{\mathbf{Z}_{:, j}} \frac{1}{2} \mathbf{Q}\mathbf{Z}_{:, j}^{\top} \mathbf{Z}_{:, j}+\boldsymbol{q}^{\top} \mathbf{Z}_{:, j} \\
				&s.t.~ \mathbf{Z} \geq 0, \mathbf{Z}_{:, j}^{\top} \mathbf{1}=1, 
			\end{split}
		\end{equation}
	\end{center}
	
	for $j=1,2,\cdots,n$, where the parameter for quadratic term is $\mathbf{Q}=2\sum_{v=1}^{p} {\alpha_v^2} \mathbf{I} $ and the first order polynomial term ${\boldsymbol{q}^\top}=-2 \sum_{v=1}^{p} {\alpha_v^2} {\mathbf{X}^{(v)}_{:,j}}^{\top}\mathbf{P} $. These are several QP(quadratic programming) problems with domain constraints. As each column of $\mathbf{Z}$ is an $m$ dimensional vector, and there are $n$ sub-problems identical to each other, the overall time cost of optimising $\mathbf{Z}$ is $\mathcal{O}(nm^3)$.

	\noindent{\textbf{Update $\boldsymbol{\alpha}$ with $\mathbf{Z}$, $\mathbf{A}$ and $\{\mathbf{W}^{(v)}_i\}_{i=1}^{\delta}$ fixed.}}
	When variables $\mathbf{Z}$, $\mathbf{A}$ and $\{\mathbf{W}^{(v)}_i\}_{i=1}^{\delta}$ are fixed, the optimization problem in Eq. \eqref{target1} with regard to the variable $\boldsymbol{\alpha}$ automatically learns the coefficients of different views, the procedure is assembled below:
	
	\begin{center}
		\begin{equation}\label{op_alpha1}
			\begin{split}
				&\mathop{\min}\limits_{\small \boldsymbol{\alpha}} \large \sum_{v=1}^{p} \alpha_{v}^{2} \| \mathbf{X}^{(v)}-\mathbf{W}^{(v)}_{1}\mathbf{W}^{(v)}_{2}\cdots\mathbf{W}^{(v)}_{\delta}\mathbf{A}\mathbf{Z}\| _{F}^2, \\
				&s.t.~ \small\boldsymbol{\alpha}^{\top} \mathbf{1}=1,\small\boldsymbol{\alpha} \geq 0,
			\end{split}
		\end{equation}
	\end{center}
	
	Denote the Frobenius norm of the alignment term $(\mathbf{X}^{(v)}-\mathbf{W}^{(v)}_{1}\mathbf{W}^{(v)}_{2}\cdots\mathbf{W}^{(v)}_{\delta}\mathbf{A}\mathbf{Z})$ as $f^{(v)}$, and the vector $\mathbf{F}=[f^{(1)},f^{(2)},\cdots,f^{(m)}]^{\top}$ refers to the concatenation. By Cauchy-Schwarz inequality, the optimal value of $\boldsymbol{\alpha}$ can be directly acquired by 
	\begin{align}
		\label{op_alpha2}
		\boldsymbol{\alpha}_i=\frac{({f}^{(i)})^{-1}}{\sum_{i=1}^p ({f}^{(i)})^{-1}}
	\end{align}

	\begin{algorithm}[htb]
		\renewcommand{\algorithmicrequire}{\textbf{Input:}}
		\renewcommand\algorithmicensure {\textbf{Output:} }
		\caption{Anchor-based Multi-view Subspace Clustering with Hierarchical Feature Descent(MVSC-HFD)
		}
		\label{Algorithm_1}
		\begin{algorithmic}[1]
			\REQUIRE ~~\\
			Multiview representations $\{ \mathbf{X}_{(1)}, \dots ,\mathbf{X}_{(p)}\}$, number of clusters $k$, depth $d$, number of anchor points $m$.
			\ENSURE ~~\\
			The shared bipartite graph $\mathbf{Z}$.\\
			\STATE \textbf{Initialization} Initialize sampling matrix $\mathbf{A} \in \mathbb{R}^{d_{\textit{anchor}} \times m}$ as an identity matrix (In our experimental setting, $d_{\textit{anchor}} = k$ and $m = k$). Set $\mathbf{Z}$ as an $m \times n$ matrix with the first $m \times m$ block deposited as identity matrix and the rest elements deposited to be $0$. Initialize the coefficient vector $\boldsymbol{\alpha}$ with the mean value factorized by the view count. Assign $t=1$.
			\REPEAT
			\STATE Calculate $\mathbf{W}^{(v)}_i(i=1,2,\cdots,\delta)$ by optimizing Eq.~\eqref{op_W_end};
			\STATE Calculate $\mathbf{A}$ by optimizing Eq.~\eqref{op_A2};
			\STATE Calculate $\mathbf{Z}$ by optimizing Eq.~\eqref{op_Z3};
			\STATE Calculate $\boldsymbol{\alpha}$ by optimizing Eq.~\eqref{op_alpha2};
			\STATE Update $t=t+1$.
			\UNTIL Convergence: {$ \| Target^{(t)} - Target^{(t-1)}\|_\mathrm{F} < 10^{-3} \times \| Target^{(t-1)}\|_\mathrm{F}$}.
		\end{algorithmic}
	\end{algorithm}
	
	\subsection{Algorithmic Discussion}
	In the section of Algorithmic Discussion, we thoroughly examine the computational complexity of our algorithm and provide a comprehensive review of our methodology.
	
	\subsubsection{Convergence Analysis} In our four-step iterative convex optimization process, each updating formula has a closed-form solution and for each iteration, the objective value of the optimization target monotonically decreases while keeping the rest of the decision variables fixed. At the same time, both $\{\mathbf{W}^{(v)}_{i}\}_{i=1}^{\delta}$ and $\mathbf{A}$ are orthogonal matrices, while both $\boldsymbol{\alpha}$ and $\mathbf{Z}$ satisfies $1$-sum non-negative constraints, which provides the lower bound of the optimization target. Consequently, the convex objective monotonically decreases with a lower bound, thus guaranteeing the overall convergence of the proposed method MVSC-HFD. 
	
	\subsubsection{Complexity Analysis} The optimization procedure demonstrated in Algorithm \ref{Algorithm_1} has decomposed the model into four sub-problems with a closed-form solution. For simplicity and practicality, the hierarchical dimensional reduction processes are conducted identically for different views. Subsequently, their layers' dimensions $\{l_i\}_{i=1}^{\delta}$ are the same. In specific, the optimization of $\mathbf{W}^{(v)}_o$ in Eq. \eqref{op_W_end} involves the calculation of $\mathbf{M}^{(v)}$ and its SVD complexity, which is $\mathcal{O}((\sum_{i=0}^{\delta-2}l_i l_{i+1})l_{\delta}+nd_vl_{\delta}+2md_vl_{\delta})$ and $\mathcal{O}(l_{o-1}k^2)$, respectively. The multiplication applied to acquire $\mathbf{W}^{(v)}_o$ is $\mathcal{O}(l_{o-1}k^2)$. In all, suppose the dimension $l_{\delta}$ of the last layer is the same as the cluster number $k$ and $\sum_{i=1}^{p}d_v=q$, it needs $\mathcal{O}((ql_1+p\sum_{i=0}^{\delta-2}l_i l_{i+1})k+nqk+2mqk)$ to update $\{\mathbf{W}^{(v)}_{i}\}_{i=1}^{\delta}$ for all views. The optimization of $\mathbf{A}$ requires $\mathcal{O}(k(d_v+m)n)$ time cost to gather the multiplicator and $\mathcal{O}(km^2)$ to decompose it. In sum, the overall complexity of our algorithm is $\mathcal{O}(n)$, which is scalable for large-scale datasets.

	\section{Experiments and Analysis}
	In the section that follows, we conduct a series of experiments to validate the effectiveness of the algorithm using synthetic and real-world datasets and compare our algorithm with state-of-the-art multi-view clustering algorithms. Last but not least, it is worthwhile to explore the mechanism of the algorithm by analyzing the experimental results. 
	\subsection{Datasets Overview and Experimental Settings}
	\subsubsection{A Brief Introduction of Datasets}
	For in-depth experimental analysis, we conduct a series of tests over 10 real-world datasets, varying in multiple domains. The datasets used in our experiments are 
	WebKB \footnote[1]{http://lig-membres.imag.fr/grimal/data.html},
	BDGP \footnote[2]{https://www.fruitfly.org/},
	MSRCV1 \cite{winn2005locus},
	Caltech7-2view \footnote[3]{http://www.vision.caltech.edu/Image\_Datasets/Caltech101/},
	NUS-WIDE-10  \footnote[4]{https://lms.comp.nus.edu.sg/wp-content/uploads/2019/research/nuswide/NUS-WIDE.html},
	Caltech101-7 $^3$,
	Caltech101-20 $^3$,
	NUS-WIDE $^4$,
	SUNRGBD \footnote[5]{https://rgbd.cs.princeton.edu/},
	YoutubeFace \footnote[6]{https://www.cs.tau.ac.il/~wolf/ytfaces/}. 
	{These datasets vary in broad ranges of application background. In specific, WebKB is a dataset that includes web pages from computer science departments of various universities and they are categorized into 5 categories (Student, Faculty, Department, Course, Project). The SUNRGBD dataset contains 10335 real RGB-D images of room scenes. The MSRCV1 dataset from Microsoft Research in Cambridge contains 7 classes of images with coarse pixel-wise labeling.}
	In Table \ref{Table:datasets1}, we summarize the detailed information of these benchmark datasets.

	\begin{table}[!t]
		\setlength{\abovecaptionskip}{0pt}
		\setlength{\belowcaptionskip}{10pt}
		\renewcommand{\arraystretch}{1.0}
		\centering
		\caption{{Information of benchmark datasets}}\label{Table:datasets1}
		\scalebox{0.95}{
			\begin{tabular}{c||c|c|c}
				\hline
				Datasets &\#  Samples &\#  Views &\# Clusters  \\
				\hline\hline
				{MSRCV1}       & 210  & 6  &  7 \\
				{WebKB}            & 265  & 2  & 5   \\
				{Caltech101-7}            & 1474 &  6  & 7   \\
				{Caltech7-2view}       & 1474  & 2  &  7 \\
				{Caltech101-20}	& 2386  & 6   &  20 \\
				{BDGP}         & 2500  & 3   &  5  \\
				{NUS-WIDE-10}       & 6251  & 5  &  10 \\
				{NUS-WIDE}       & 23953  & 5  &  31 \\
				{SUNRGBD}       & 10335  & 2  &  45 \\
				{YoutubeFace}       & 101499  & 5  &  31 \\
				\hline
		\end{tabular}}
	\end{table}

	\subsubsection{Experimental Settings}
	We use the initialization specified in Algorithm \ref{Algorithm_1}. In practice, we simply adopt the number of clusters $k$ as the value of the dimension of embedding space and the value of anchor count. For each view, the hierarchical feature descent rate is different according to their original dimensions. However, it is rather practicable as it performs well by directly slicing the dimension to equal sections: $l_o-l_{o-1} = l_{o+1}-l_{o}, \forall o \in [1,\delta]$. {The best depths among $[1,2,\cdots,5]$ of the datasets WebKB, BDGP, MSRCV1, Caltech7-2view,
		NUS-WIDE-10, Caltech101-7, Caltech101-20, NUS-WIDE $^4$, SUNRGBD, YoutubeFace are 5, 5, 5, 5, 4, 4, 5, 5, 2, respectively.} 
	
	{To be specific, the experiments are conducted in MATLAB environment on PC with Intel(R) Core(TM) i7-6800K 3.40GHz CPU and 128GB RAM.}
	
	\subsubsection{Compared Algorithms}
	We list the information of our compared algorithms as in detail as possible by first giving brief introduction to their distinct mechanism, and analyze the hyper-parameters of each method.
	\begin{itemize}
		\item \textbf{Multi-view k-means clustering on big data (RMKM)~\cite{cai2013multi}.} This method uses $\ell_{2,1}$ norm to minimize the non-negative matrix factorization form of the cluster indicator matrix in terms of reconstruction of the original data matrix. Its inherent property is similar to the $k$-means clustering method and the $1$-$\text{of}$-$K$ coding scheme provides one-step clustering results.
		\item \textbf{Large-scale multi-view subspace clustering in linear time (LM-\\VSC)~\cite{kang2020large}.} 
		Under the framework of subspace clustering, this paper intends for clustering of large-scale datasets by using initially selected anchor points to build reconstruction. 
		\item \textbf{Parameter-free auto-weighted multiple graph learning: a fram-ework for multi-view clustering and semi-supervised classification (AMGL)~\cite{nie2016parameter}.} A fusion of different graph representations where the weights of each graph of view are automatically learned and assumed to be the optimal combination.
		\item \textbf{Binary Multi-view Clustering (BMVC)~\cite{zhang2018binary}.} In order to reduce computational complexity and spacial cost, this work proposes to integrate binary representation of all views and consensus binary clustering into a uniform framework by both minimizing these two parts of loss at the same time.
		\item \textbf{Flexible Multi-View Representation Learning for Subspace Clustering (FMR)~\cite{li2019flexible}.} After performing clustering on each different view, the best clustering result is flexibly encoded via adopting complementary information of provided views, thus avoiding the partial outcome in the reconstruction stage. 
		\item \textbf{Multi-view Subspace Clustering (MVSC) \cite{gao2015multi}.} MVSC proposes to learn subspace representation of different views simultaneously while regularising the separate affinity matrices to a common one to improve clustering consistency among different views. 
		\item \textbf{Multi-view clustering: A scalable and parameter-free bipartite graph fusion method (SFMC)~\cite{li2020multi}.} Based on Ky Fan’s Theorem~\cite{fan1949theorem}, SFMC explores the relationship between Laplacian matrix and graph connectivity, thus proposing a scalable and parameter-free method to interactively fuse the view-specific graphs into a consensus graph with adaptable anchor strategy. 
		\item \textbf{Fast parameter-free multi-view subspace clustering with consensus anchor guidance (FPMVS-CAG)~\cite{wang2021fast}.}
		FPMVS-CAG integrates anchor guidance and bipartite graph construction within an optimization framework and learns collaboratively, promoting clustering quality. The algorithm offers linear time complexity, is hyper-parameter-free, and automatically learns an optimal anchor subspace graph.
		\item \textbf{Efficient Multi-view K-means Clustering with Multiple Anchor Graphs (EMKMC)~\cite{yang2022efficient}}.
		EMKMC integrates anchor graph-based and $k$-means-based approaches, constructing anchor graphs for each view and utilizing an improved $k$-means strategy for integration. This enables high efficiency, particularly for large-scale high-dimensional multi-view data, while maintaining or surpassing clustering effectiveness compared to other methods.
	\end{itemize}
	
	\begin{table}[]
		\caption{{Parameter analysis of the compared algorithms.}}
		\resizebox{0.99\textwidth}{!}{
			\begin{tabular}{c|c|c|c|c|c|c|c|c|c|c}
				\hline 
				& RMKM & LMVSC & AMGL & BMVC & FMR & MVSC & SFMC & FPMVS-CAG & EMKMC & {proposed}\tabularnewline
				\hline 
				\hline 
				\#Parameter & 1 & 2 & 0 & 3 & 2 & 3 & 1 & 0 & $p$ & 1 \tabularnewline
				\hline 
		\end{tabular}}
		\label{tab:parameter}
	\end{table}
	
	As shown in Table.\ref{tab:parameter}, most of algorithm have more than one hyper-parameters. It is encouraged to use learnable parameters which can adaptively collaborate with the optimization of other decision variables to boost the performance whereas spare the acquisition of prior knowledge. {In our method, the learnable vector of parameters $\boldsymbol{\alpha}=[\alpha_1,\alpha_2,\cdots,\alpha_p]^{\top}$ represents the weights adherent to different views, and they are optimized in each iteration.} However, {for the compared algorithm EMKMC, each view has a hyper-parameter, so the number of parameters is equal to the number of views, which we denote by $p$. As a result, the searching space for best parameter is $\epsilon^p$ where $\epsilon$ denotes the size of the candidate set of the hyper-parameter. These unlearnable parameters dramatically increases computation time of EMKMC.}
	
	\subsection{Comparison With State-of-the-Art Algorithms}
	In this section, we conduct a comparative analysis between our algorithm and several state-of-the-art MVC algorithms to validate its effectiveness. We evaluate the performance of our algorithm using 9 widely-used methods, considering accuracy (ACC), Normalized Mutual Information (NMI), and purity as evaluation metrics. {The results are presented in Table \ref{tab:all-res}}. It shows that our findings entail most of the best practices and recommendations that can improve multi-view learning performance. 
	
\begin{figure}[H]
		\centering
		\subfloat[\scriptsize{WebKB}]{\includegraphics[width = 0.33\textwidth]{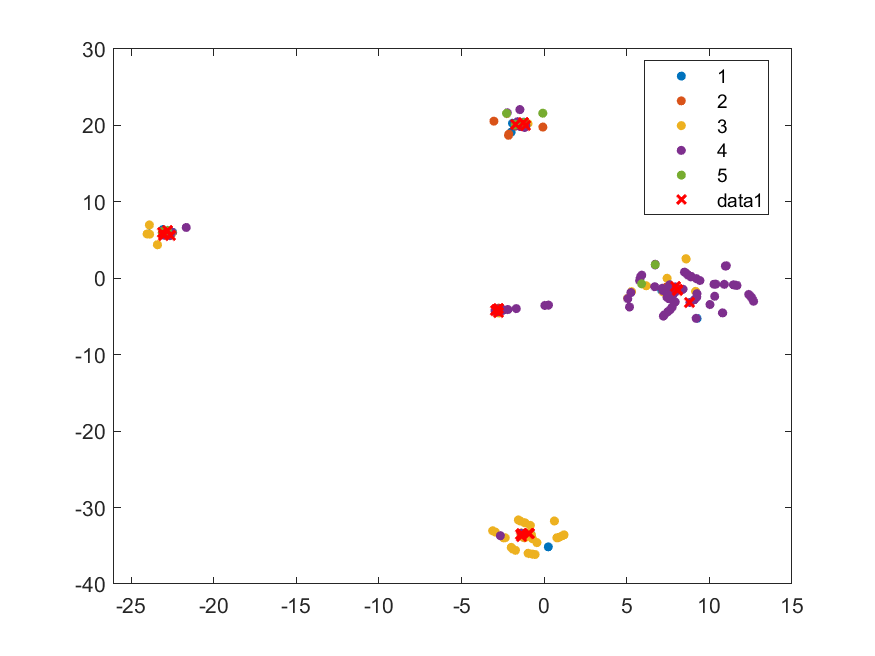}}%
		\subfloat[\scriptsize{MSRCV1}]{\includegraphics[width = 0.33\textwidth]{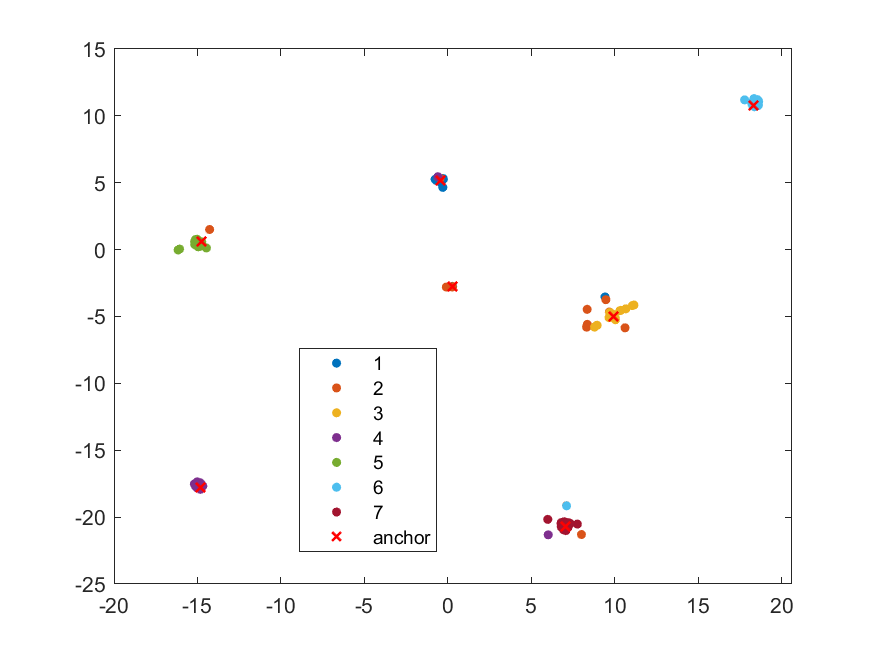}}
		\subfloat[\scriptsize{Caltech7-2views}]{\includegraphics[width = 0.33\textwidth]{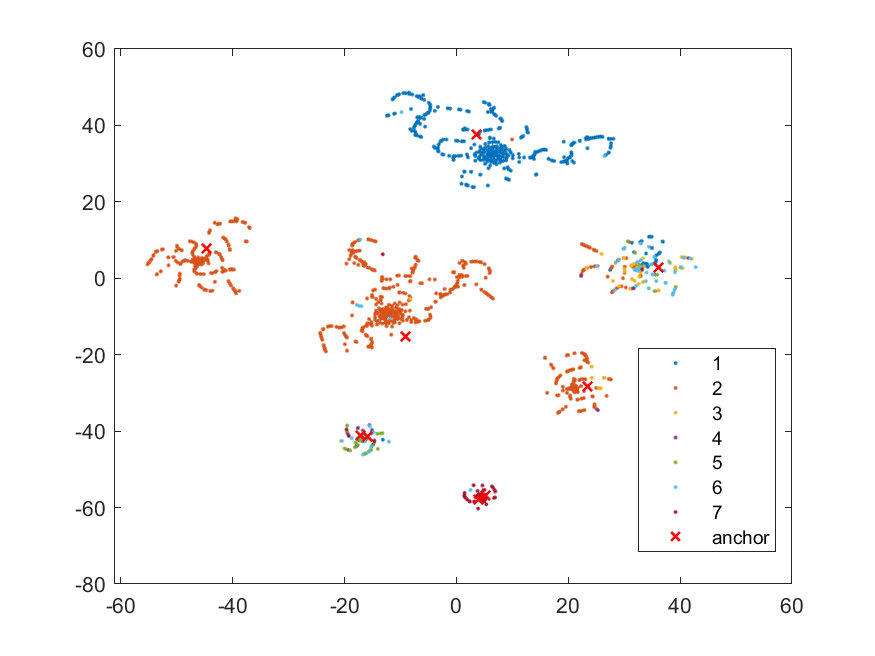}}
		\caption{{The t-SNE plot of our proposed method on 3 benchmark datasets}}
		\label{fig:tsne}
	\end{figure}

	{To illustrate the effectiveness of the anchors, we conduct t-Distributed Stochastic Neighbor Embedding (t-SNE) experiment and mark the anchors with red crosses. Due to space constraints, we show the clustering effect for WebKB, MSRCV1 and Caltech7-2views in Figure \ref{fig:tsne}. We observe that a red cross is located in each cluster, which empirically verifies the effectiveness of the optimization of anchors in the objective function.}
	

	\begin{table}[]
		\caption{{Comparison of clustering performance and experimental results on benchmark datasets, analyzed with baseline methods. Slash(-) means out-of-memory failure.}}
		\resizebox{0.99\textwidth}{!}{
			\begin{tabular}{@{}lcccccccccc@{}}
				\toprule
				& RMKM & LMVSC & AMGL & BMVC & FMR & MVSC & SFMC & FPMVS-CAG & EMKMC & Proposed \\ \midrule
				\multicolumn{11}{c}{ACC} \\ \midrule
				WebKB & 0.4151 & {\ul 0.5472} & 0.2264 & 0.3925 & 0.4943 & 0.4830 & 0.4755 & 0.4830 & 0.4455 & \textbf{0.6189} \\
				BDGP & 0.4144 & {\ul 0.4540} & 0.3328 & 0.3292 & 0.3860 & 0.3600 & 0.2012 & 0.3892 & 0.3186 & \textbf{0.4796} \\
				MSRCV1 & 0.5429 & 0.7905 & 0.7381 & 0.3619 & 0.8381 & 0.4571 & 0.7000 & {\ul 0.8714} & 0.4027 & \textbf{0.8952} \\
				Caltech7-2view & 0.4233 & \textbf{0.7022} & 0.3867 & 0.4417 & 0.5115 & 0.4423 & 0.6737 & 0.5916 & 0.4453 & {\ul 0.7008} \\
				NUS-WIDE-10 & 0.1992 & 0.2342 & 0.1339 & {\ul 0.2544} & 0.2492 & 0.1899 & 0.2174 & 0.2464 & 0.1956 & \textbf{0.2553} \\
				Caltech101-7 & 0.2877 & 0.6248 & 0.3996 & 0.3874 & 0.4552 & 0.5204 & 0.6526 & {\ul 0.6988} & 0.4632 & \textbf{0.7205} \\
				Caltech101-20 & 0.3345 & 0.4149 & 0.2993 & 0.4011 & 0.3822 & 0.3651 & {\ul 0.5256} & 0.5163 & 0.3108 & \textbf{0.5298} \\
				NUS-WIDE & 0.1336 & 0.1298 & 0.0493 & 0.1461 & - & - & 0.1341 & {\ul 0.1680} & 0.0940 & \textbf{0.1761} \\
				SUNRGBD & 0.1836 & 0.1628 & 0.1002 & 0.1538 & - & - & 0.1136 & {\ul 0.2022} & 0.1164 & \textbf{0.2222} \\
				YoutubeFace & - & \textbf{0.1471} & - & 0.0904 & - & - & - & 0.1359 & 0.0903 & {\ul 0.1380} \\ \midrule
				\multicolumn{11}{c}{NMI} \\ \midrule
				WebKB & 0.1068 & \textbf{0.3175} & 0.0464 & 0.0960 & 0.1266 & 0.0761 & 0.0476 & 0.1361 & 0.1868 & {\ul 0.2566} \\
				BDGP & \textbf{0.2812} & {\ul 0.2567} & 0.1460 & 0.0848 & 0.1090 & 0.1063 & 0.0032 & 0.1531 & 0.0840 & 0.2450 \\
				MSRCV1 & 0.4850 & 0.7123 & 0.7639 & 0.1955 & 0.7330 & 0.3627 & 0.7197 & {\ul 0.8151} & 0.2946 & \textbf{0.8420} \\
				Caltech7-2view & 0.4981 & 0.5253 & 0.4252 & 0.4139 & 0.4443 & 0.3268 &  0.5862 & {\ul 0.5959} & 0.2930 & \textbf{0.5989} \\
				NUS-WIDE-10 & 0.0789 & 0.1076 & 0.0107 & \textbf{0.1531} & {\ul 0.1169} & 0.0606 & 0.0176 & 0.1088 & 0.0686 & 0.1111 \\
				Caltech101-7 & 0.1411 & 0.5345 & 0.4514 & {\ul 0.5579} & 0.3225 & 0.3823 & \textbf{0.5629} & 0.4580 & 0.3007 & 0.4949 \\
				Caltech101-20 & 0.0121 & 0.5125 & 0.4920 & \textbf{0.5527} & {\ul 0.5323} & 0.4495 & 0.4942 & 0.4837 & 0.3316 & 0.5169 \\
				NUS-WIDE & 0.0016 & 0.1041 & 0.0140 & \textbf{0.1510} & - & - & 0.0049 & 0.1228 & 0.0581 & {\ul 0.1240} \\
				SUNRGBD & \textbf{0.2343} & {\ul 0.2321} & 0.1857 & 0.1758 & - & - & 0.0230 & 0.2104 & 0.1158 & 0.2245 \\
				YoutubeFace & - & \textbf{0.1319} & - & 0.0568 & - & - & - & 0.1039 & 0.0616 & {\ul 0.1050} \\ \midrule
				\multicolumn{11}{c}{Purity} \\ \midrule
				WebKB & 0.5321 & \textbf{0.6943} & 0.2340 & 0.5208 & 0.5660 & 0.5019 & 0.4830 & 0.5358 & 0.5878 & {\ul 0.6415} \\
				BDGP & {\ul 0.4576} & 0.4540 & 0.3504 & 0.3592 & 0.3860 & 0.3600 & 0.2016 & 0.4044 & 0.3300 & \textbf{0.4852} \\
				MSRCV1 & 0.5810 & 0.7905 & 0.8000 & 0.3810 & 0.8381 & 0.4762 & 0.7238 & {\ul 0.8714} & 0.4304 & \textbf{0.8952} \\
				Caltech7-2view & 0.8589 & 0.8514 & 0.3962 & 0.7910 & 0.8318 & 0.7802 & {\ul 0.8636} & 0.8632 & 0.7621 & \textbf{0.8711} \\
				NUS-WIDE-10 & 0.3172 & 0.3232 & 0.1366 & \textbf{0.3745} & {\ul 0.3620} & 0.2788 & 0.2284 & 0.3120 & 0.3059 & 0.3158 \\
				Caltech101-7 & 0.6608 & {\ul 0.8541} & 0.3996 & \textbf{0.8548} & 0.7096 & 0.8033 & 0.8528 & 0.8033 & 0.7689 & 0.8297 \\
				Caltech101-20 & 0.3345 & 0.6903 & 0.3215 & \textbf{0.7334} & {\ul 0.7091} & 0.6568 & 0.6249 & 0.6622 & 0.5625 & 0.6811 \\
				NUS-WIDE & 0.1336 & 0.2148 & 0.0502 & \textbf{0.2622} & - & - & 0.1353 & 0.2238 & 0.1859 & {\ul 0.2240} \\
				SUNRGBD & \textbf{0.3771} & {\ul 0.3536} & 0.1098 & 0.2984 & - & - & 0.1163 & 0.2999 & 0.2470 & 0.3111 \\
				YoutubeFace & - & \textbf{0.2886} & - & 0.2675 & - & - & - & 0.2743 & 0.2664 & {\ul 0.2766} \\ \bottomrule
		\end{tabular}}
		\label{tab:all-res}
	\end{table}

	
	\subsubsection{Running Time Comparison}
\begin{figure}[h]
		\centering
		\includegraphics[width=0.99\textwidth]{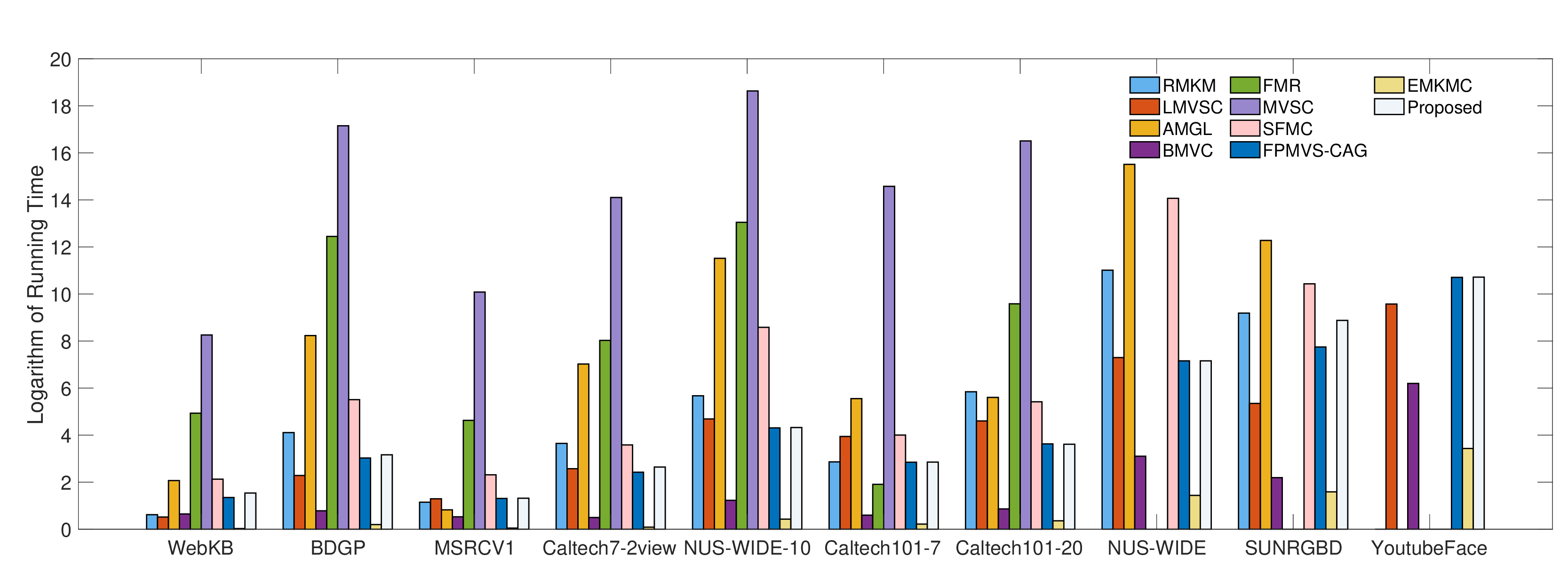}
		\caption{The relative running time of compared methods.}
		\label{fig:time}
	\end{figure}

	The running time of compared algorithms on benchmark datasets are shown in Figure \ref{fig:time}. {Due to the excessive time consumed by some of the algorithms, we use the logarithms of the time as the vertical axis of the coordinate.}
	
	It is worth noticing that for large-scale datasets, namely NUS-WIDE, SUNRGBD and YoutubeFace, some algorithms(FMR, MVSC, SFMC, RMKM) encounter the 'Out-of-Memory' problem. As a result, the time bars of these methods are omitted in the comparison graph. {We observe that the time cost of EMKMC is comparable to ours, but the total runtime cost is still high owing to the aforementioned search space for the best hyper-parameters.}
	
	\subsubsection{Convergence}
\begin{figure}[H]
		\centering
		\subfloat[\scriptsize{WebKB}]{\includegraphics[width = 0.20\textwidth]{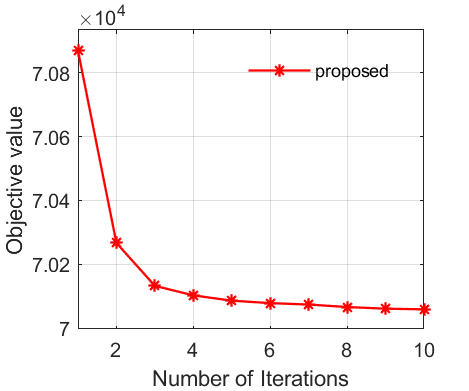}}%
		\subfloat[\scriptsize{BDGP}]{\includegraphics[width = 0.20\textwidth]{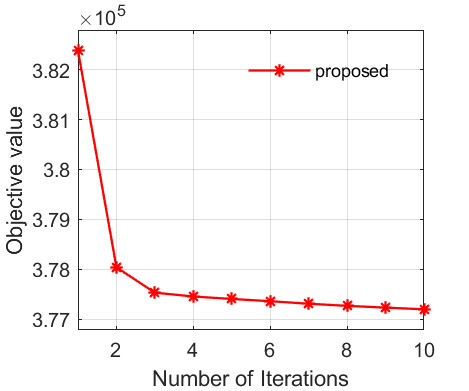}}
		\subfloat[\scriptsize{MSRCV1}]{\includegraphics[width = 0.20\textwidth]{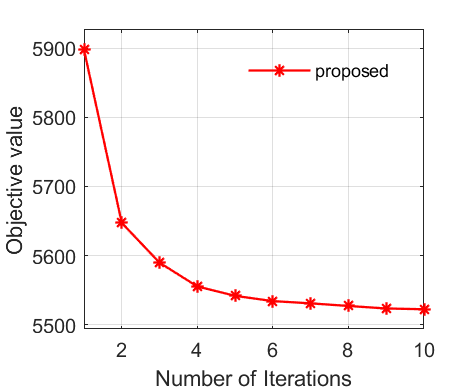}}
		\subfloat[\scriptsize{Caltech7-2view}]{\includegraphics[width = 0.20\textwidth]{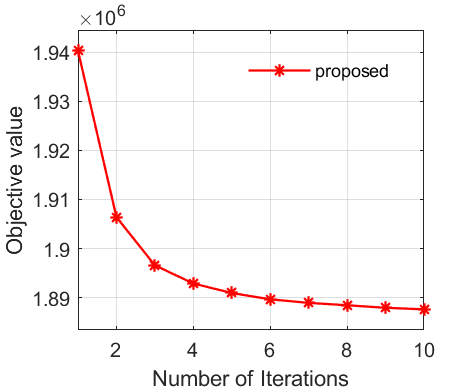}}
		\subfloat[\scriptsize{NUS-WIDE-10}]{\includegraphics[width = 0.20\textwidth]{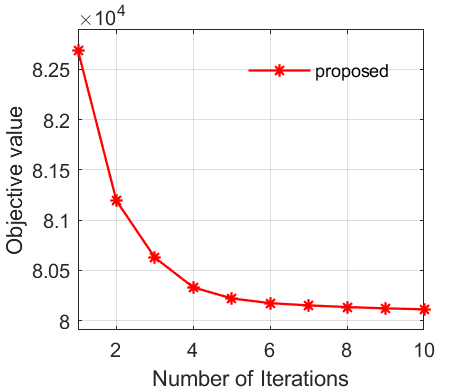}}
		
		\subfloat[\scriptsize{Caltech101-7}]{\includegraphics[width = 0.20\textwidth]{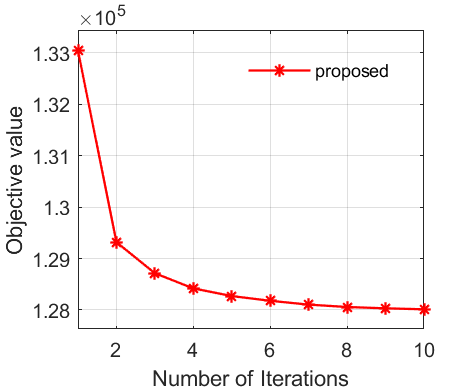}}
		\subfloat[\scriptsize{Caltech101-20}]{\includegraphics[width = 0.20\textwidth]{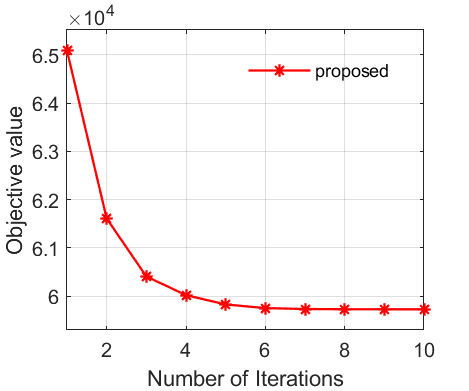}}
		\subfloat[\scriptsize{NUS-WIDE}]{\includegraphics[width = 0.20\textwidth]{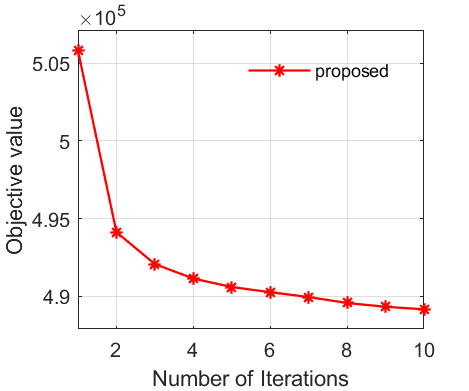}}
		\subfloat[\scriptsize{SUNRGBD}]{\includegraphics[width = 0.20\textwidth]{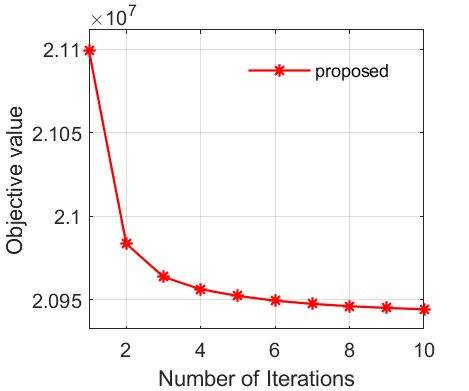}}
		\subfloat[\scriptsize{YoutubeFace}]{\includegraphics[width = 0.20\textwidth]{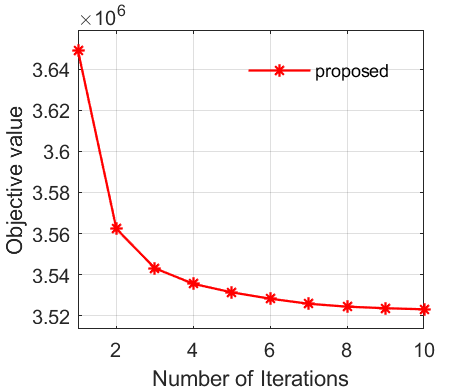}}
		\caption{The objective curve of our proposed method on 10 benchmark datasets}
		\label{fig:objective}
	\end{figure}

	As described in the previous sections, our proposed algorithm possesses a theoretical guarantee, ensuring its convergence to a local minimum. 
	Furthermore, to provide empirical evidence of its convergence, we conduct experiments on benchmark datasets. Figure \ref{fig:objective} illustrates the evolution of the objective value throughout the experimental process. Notably, the experimental results reveal a consistent and monotonically decreasing trend in the objective values of our algorithm at each iteration. Furthermore, we observe that the algorithm achieves rapid convergence, typically within fewer than 10 iterations. These findings undeniably validate the convergence property of our proposed algorithm.
	
	\subsubsection{Parameter Analysis} 
\begin{figure}[h]
		\centering
		\includegraphics[width=1\textwidth]{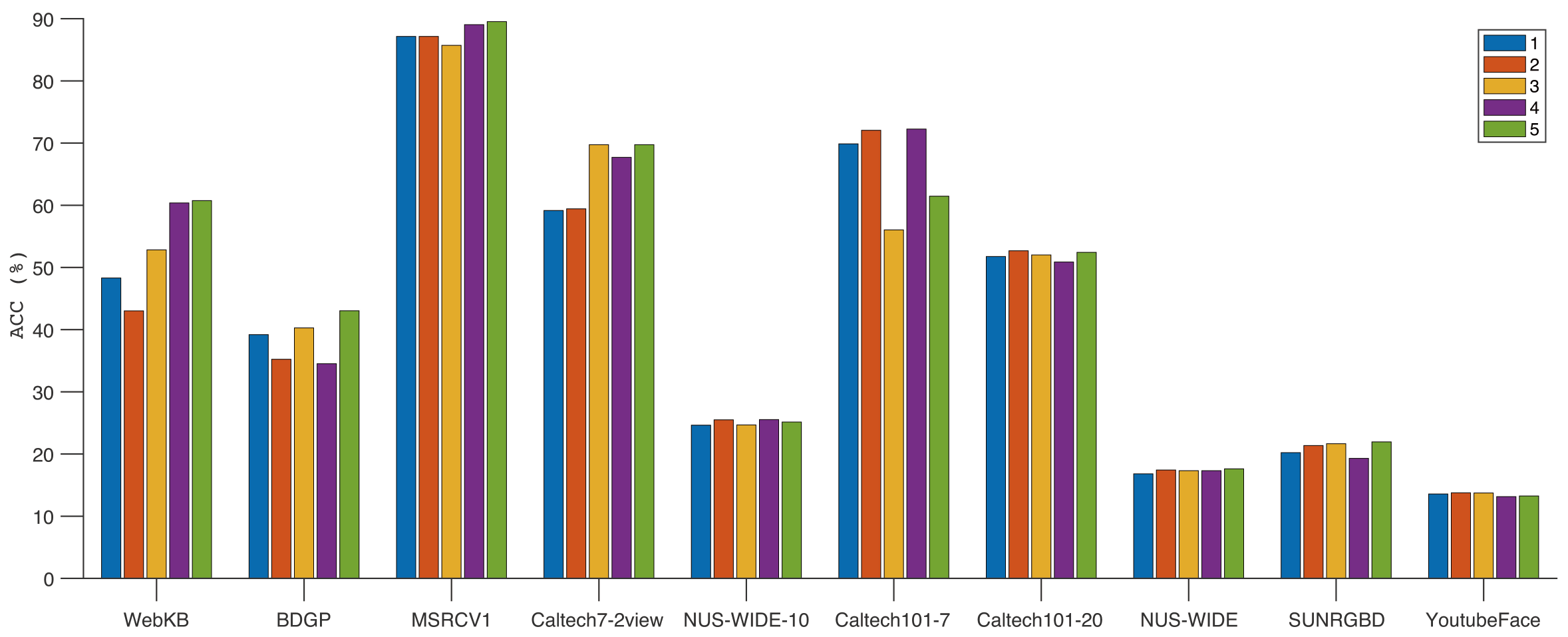}
		\caption{The sensitivity of layers of the proposed method.}
		\label{fig:sen}
	\end{figure}

	To demonstrate the impact of varying selections for the number of hierarchical dimension projection matrix, we present the performance variation across distinct layers of $\mathbf{W}^{(v)}_i$ with respect to each dataset, as depicted in Figure \ref{fig:sen}.
	
	{
		\subsection{Conclusion}
		Our work focuses on multi-view clustering and explore ways to increase the effectiveness of anchor-based subspace clustering by tackling the discrepancy and deploying the dependency of different views. We will further investigate whether the number of categories of the datasets can positively affect the accuracy of clustering by providing guidance for the assignation of the number of anchors; and whether it can be beneficial to the heuristic setting of the dimensions of each layer. }

	
	
	\bibliographystyle{elsarticle-harv} 
	\bibliography{ref}

\begin{thebibliography}{51}
\expandafter\ifx\csname natexlab\endcsname\relax\def\natexlab#1{#1}\fi
\providecommand{\url}[1]{\texttt{#1}}
\providecommand{\href}[2]{#2}
\providecommand{\path}[1]{#1}
\providecommand{\DOIprefix}{doi:}
\providecommand{\ArXivprefix}{arXiv:}
\providecommand{\URLprefix}{URL: }
\providecommand{\Pubmedprefix}{pmid:}
\providecommand{\doi}[1]{\href{http://dx.doi.org/#1}{\path{#1}}}
\providecommand{\Pubmed}[1]{\href{pmid:#1}{\path{#1}}}
\providecommand{\bibinfo}[2]{#2}
\ifx\xfnm\relax \def\xfnm[#1]{\unskip,\space#1}\fi
\bibitem[{Akaho(2006)}]{DBLP:journals/corr/abs-cs-0609071}
\bibinfo{author}{Akaho, S.}, \bibinfo{year}{2006}.
\newblock \bibinfo{title}{A kernel method for canonical correlation analysis}.
\newblock \bibinfo{journal}{CoRR} \bibinfo{volume}{abs/cs/0609071}.
\bibitem[{Andrew et~al.(2013)Andrew, Arora, Bilmes and
  Livescu}]{DBLP:conf/icml/AndrewABL13}
\bibinfo{author}{Andrew, G.}, \bibinfo{author}{Arora, R.},
  \bibinfo{author}{Bilmes, J.A.}, \bibinfo{author}{Livescu, K.},
  \bibinfo{year}{2013}.
\newblock \bibinfo{title}{Deep canonical correlation analysis}, in:
  \bibinfo{booktitle}{Proceedings of the 30th International Conference on
  Machine Learning, {ICML} 2013, Atlanta, GA, USA, 16-21 June 2013},
  \bibinfo{publisher}{JMLR.org}. pp. \bibinfo{pages}{1247--1255}.
\bibitem[{Bellman(1954)}]{DBLP:journals/ior/Bellman54}
\bibinfo{author}{Bellman, R.}, \bibinfo{year}{1954}.
\newblock \bibinfo{title}{Some applications of the theory of dynamic
  programming - {A} review}.
\newblock \bibinfo{journal}{Oper. Res.} , \bibinfo{pages}{275--288}.
\bibitem[{Cai et~al.(2023)Cai, Huang, Zhang and Wang}]{cai2023seeking}
\bibinfo{author}{Cai, X.}, \bibinfo{author}{Huang, D.}, \bibinfo{author}{Zhang,
  G.Y.}, \bibinfo{author}{Wang, C.D.}, \bibinfo{year}{2023}.
\newblock \bibinfo{title}{Seeking commonness and inconsistencies: A jointly
  smoothed approach to multi-view subspace clustering}.
\newblock \bibinfo{journal}{Information Fusion} \bibinfo{volume}{91},
  \bibinfo{pages}{364--375}.
\bibitem[{Cai et~al.(2013)Cai, Nie and Huang}]{cai2013multi}
\bibinfo{author}{Cai, X.}, \bibinfo{author}{Nie, F.}, \bibinfo{author}{Huang,
  H.}, \bibinfo{year}{2013}.
\newblock \bibinfo{title}{Multi-view k-means clustering on big data}, in:
  \bibinfo{booktitle}{Twenty-Third International Joint conference on artificial
  intelligence}.
\bibitem[{Cao et~al.(2015)Cao, Zhang, Fu, Liu and
  Zhang}]{DBLP:conf/cvpr/CaoZFLZ15}
\bibinfo{author}{Cao, X.}, \bibinfo{author}{Zhang, C.}, \bibinfo{author}{Fu,
  H.}, \bibinfo{author}{Liu, S.}, \bibinfo{author}{Zhang, H.},
  \bibinfo{year}{2015}.
\newblock \bibinfo{title}{Diversity-induced multi-view subspace clustering},
  in: \bibinfo{booktitle}{{IEEE} Conference on Computer Vision and Pattern
  Recognition, {CVPR} 2015, Boston, MA, USA, June 7-12, 2015}.
\bibitem[{Chen and Cai(2011)}]{chen2011large}
\bibinfo{author}{Chen, X.}, \bibinfo{author}{Cai, D.}, \bibinfo{year}{2011}.
\newblock \bibinfo{title}{Large scale spectral clustering with landmark-based
  representation}, in: \bibinfo{booktitle}{Twenty-fifth AAAI conference on
  artificial intelligence}.
\bibitem[{Chen et~al.(2022)Chen, Lin, Chen, Ye and Wang}]{chen2022diversity}
\bibinfo{author}{Chen, Z.}, \bibinfo{author}{Lin, P.}, \bibinfo{author}{Chen,
  Z.}, \bibinfo{author}{Ye, D.}, \bibinfo{author}{Wang, S.},
  \bibinfo{year}{2022}.
\newblock \bibinfo{title}{Diversity embedding deep matrix factorization for
  multi-view clustering}.
\newblock \bibinfo{journal}{Information Sciences} \bibinfo{volume}{610},
  \bibinfo{pages}{114--125}.
\bibitem[{Elhamifar and Vidal(2013)}]{DBLP:journals/pami/ElhamifarV13}
\bibinfo{author}{Elhamifar, E.}, \bibinfo{author}{Vidal, R.},
  \bibinfo{year}{2013}.
\newblock \bibinfo{title}{Sparse subspace clustering: Algorithm, theory, and
  applications}.
\newblock \bibinfo{journal}{{IEEE} Trans. Pattern Anal. Mach. Intell.} .
\bibitem[{Fan(1949)}]{fan1949theorem}
\bibinfo{author}{Fan, K.}, \bibinfo{year}{1949}.
\newblock \bibinfo{title}{On a theorem of weyl concerning eigenvalues of linear
  transformations i}.
\newblock \bibinfo{journal}{Proceedings of the National Academy of Sciences}
  \bibinfo{volume}{35}, \bibinfo{pages}{652--655}.
\bibitem[{Gao et~al.(2015a)Gao, Nie, Li and Huang}]{DBLP:conf/iccv/GaoNLH15}
\bibinfo{author}{Gao, H.}, \bibinfo{author}{Nie, F.}, \bibinfo{author}{Li, X.},
  \bibinfo{author}{Huang, H.}, \bibinfo{year}{2015}a.
\newblock \bibinfo{title}{Multi-view subspace clustering}, in:
  \bibinfo{booktitle}{2015 {IEEE} International Conference on Computer Vision,
  {ICCV} 2015, Santiago, Chile, December 7-13, 2015}.
\bibitem[{Gao et~al.(2015b)Gao, Nie, Li and Huang}]{gao2015multi}
\bibinfo{author}{Gao, H.}, \bibinfo{author}{Nie, F.}, \bibinfo{author}{Li, X.},
  \bibinfo{author}{Huang, H.}, \bibinfo{year}{2015}b.
\newblock \bibinfo{title}{Multi-view subspace clustering}, in:
  \bibinfo{booktitle}{Proceedings of the IEEE international conference on
  computer vision}, pp. \bibinfo{pages}{4238--4246}.
\bibitem[{Guo(2013)}]{DBLP:conf/aaai/Guo13}
\bibinfo{author}{Guo, Y.}, \bibinfo{year}{2013}.
\newblock \bibinfo{title}{Convex subspace representation learning from
  multi-view data}, in: \bibinfo{booktitle}{Proceedings of the Twenty-Seventh
  {AAAI} Conference on Artificial Intelligence, July 14-18, 2013, Bellevue,
  Washington, {USA}}, \bibinfo{publisher}{{AAAI} Press}.
\bibitem[{Hotelling(1992)}]{hotelling1992relations}
\bibinfo{author}{Hotelling, H.}, \bibinfo{year}{1992}.
\newblock \bibinfo{title}{Relations between two sets of variates}, in:
  \bibinfo{booktitle}{Breakthroughs in statistics}.
  \bibinfo{publisher}{Springer}, pp. \bibinfo{pages}{162--190}.
\bibitem[{Huang et~al.(2021)Huang, Tsang, Xu, Lv and
  Liu}]{DBLP:conf/mm/HuangTX0L21}
\bibinfo{author}{Huang, S.}, \bibinfo{author}{Tsang, I.W.},
  \bibinfo{author}{Xu, Z.}, \bibinfo{author}{Lv, J.}, \bibinfo{author}{Liu,
  Q.}, \bibinfo{year}{2021}.
\newblock \bibinfo{title}{{CDD:} multi-view subspace clustering via cross-view
  diversity detection}, in: \bibinfo{booktitle}{{MM} '21: {ACM} Multimedia
  Conference, Virtual Event, China, October 20 - 24, 2021},
  \bibinfo{publisher}{{ACM}}. pp. \bibinfo{pages}{2308--2316}.
\bibitem[{Kang et~al.(2020)Kang, Zhou, Zhao, Shao, Han and Xu}]{kang2020large}
\bibinfo{author}{Kang, Z.}, \bibinfo{author}{Zhou, W.}, \bibinfo{author}{Zhao,
  Z.}, \bibinfo{author}{Shao, J.}, \bibinfo{author}{Han, M.},
  \bibinfo{author}{Xu, Z.}, \bibinfo{year}{2020}.
\newblock \bibinfo{title}{Large-scale multi-view subspace clustering in linear
  time}, in: \bibinfo{booktitle}{Proceedings of the AAAI Conference on
  Artificial Intelligence}, pp. \bibinfo{pages}{4412--4419}.
\bibitem[{Li et~al.(2019)Li, Zhang, Hu, Zhu and Wang}]{li2019flexible}
\bibinfo{author}{Li, R.}, \bibinfo{author}{Zhang, C.}, \bibinfo{author}{Hu,
  Q.}, \bibinfo{author}{Zhu, P.}, \bibinfo{author}{Wang, Z.},
  \bibinfo{year}{2019}.
\newblock \bibinfo{title}{Flexible multi-view representation learning for
  subspace clustering.}, in: \bibinfo{booktitle}{IJCAI}, pp.
  \bibinfo{pages}{2916--2922}.
\bibitem[{Li et~al.(2020)Li, Zhang, Wang and Nie}]{li2020multi}
\bibinfo{author}{Li, X.}, \bibinfo{author}{Zhang, H.}, \bibinfo{author}{Wang,
  R.}, \bibinfo{author}{Nie, F.}, \bibinfo{year}{2020}.
\newblock \bibinfo{title}{Multi-view clustering: A scalable and parameter-free
  bipartite graph fusion method}.
\newblock \bibinfo{journal}{IEEE Transactions on Pattern Analysis and Machine
  Intelligence} .
\bibitem[{Li et~al.(2015)Li, Nie, Huang and Huang}]{DBLP:conf/aaai/LiNHH15}
\bibinfo{author}{Li, Y.}, \bibinfo{author}{Nie, F.}, \bibinfo{author}{Huang,
  H.}, \bibinfo{author}{Huang, J.}, \bibinfo{year}{2015}.
\newblock \bibinfo{title}{Large-scale multi-view spectral clustering via
  bipartite graph}, in: \bibinfo{booktitle}{Proceedings of the Twenty-Ninth
  {AAAI} Conference on Artificial Intelligence, January 25-30, 2015},
  \bibinfo{publisher}{{AAAI} Press}. pp. \bibinfo{pages}{2750--2756}.
\bibitem[{Liang et~al.(2022)Liang, Liu, Zhou, Liu, Wang and
  Zhu}]{liang2022robust}
\bibinfo{author}{Liang, W.}, \bibinfo{author}{Liu, X.}, \bibinfo{author}{Zhou,
  S.}, \bibinfo{author}{Liu, J.}, \bibinfo{author}{Wang, S.},
  \bibinfo{author}{Zhu, E.}, \bibinfo{year}{2022}.
\newblock \bibinfo{title}{Robust graph-based multi-view clustering}, in:
  \bibinfo{booktitle}{Proceedings of the AAAI Conference on Artificial
  Intelligence}, pp. \bibinfo{pages}{7462--7469}.
\bibitem[{Liu et~al.(2013)Liu, Lin, Yan, Sun, Yu and
  Ma}]{DBLP:journals/pami/LiuLYSYM13}
\bibinfo{author}{Liu, G.}, \bibinfo{author}{Lin, Z.}, \bibinfo{author}{Yan,
  S.}, \bibinfo{author}{Sun, J.}, \bibinfo{author}{Yu, Y.},
  \bibinfo{author}{Ma, Y.}, \bibinfo{year}{2013}.
\newblock \bibinfo{title}{Robust recovery of subspace structures by low-rank
  representation}.
\newblock \bibinfo{journal}{{IEEE} Trans. Pattern Anal. Mach. Intell.} .
\bibitem[{Nie et~al.(2016)Nie, Li, Li et~al.}]{nie2016parameter}
\bibinfo{author}{Nie, F.}, \bibinfo{author}{Li, J.}, \bibinfo{author}{Li, X.},
  et~al., \bibinfo{year}{2016}.
\newblock \bibinfo{title}{Parameter-free auto-weighted multiple graph learning:
  a framework for multiview clustering and semi-supervised classification.},
  in: \bibinfo{booktitle}{IJCAI}, pp. \bibinfo{pages}{1881--1887}.
\bibitem[{Ou et~al.(2020)Ou, Wang, Zhou, Li, Guo and Zhu}]{ou2020anchor}
\bibinfo{author}{Ou, Q.}, \bibinfo{author}{Wang, S.}, \bibinfo{author}{Zhou,
  S.}, \bibinfo{author}{Li, M.}, \bibinfo{author}{Guo, X.},
  \bibinfo{author}{Zhu, E.}, \bibinfo{year}{2020}.
\newblock \bibinfo{title}{Anchor-based multiview subspace clustering with
  diversity regularization}.
\newblock \bibinfo{journal}{IEEE MultiMedia} \bibinfo{volume}{27},
  \bibinfo{pages}{91--101}.
\bibitem[{Qaqish et~al.(2017)Qaqish, O'Brien, Hibbard and
  Clowers}]{DBLP:journals/bioinformatics/QaqishOHC17}
\bibinfo{author}{Qaqish, B.F.}, \bibinfo{author}{O'Brien, J.J.},
  \bibinfo{author}{Hibbard, J.C.}, \bibinfo{author}{Clowers, K.J.},
  \bibinfo{year}{2017}.
\newblock \bibinfo{title}{Accelerating high-dimensional clustering with
  lossless data reduction}.
\newblock \bibinfo{journal}{Bioinform.} \bibinfo{volume}{33},
  \bibinfo{pages}{2867--2872}.
\bibitem[{Shakhnarovich(2005)}]{DBLP:phd/ndltd/Shakhnarovich05}
\bibinfo{author}{Shakhnarovich, G.}, \bibinfo{year}{2005}.
\newblock \bibinfo{title}{Learning task-specific similarity}.
\newblock Ph.D. thesis. Massachusetts Institute of Technology, Cambridge, MA,
  {USA}.
\bibitem[{Shu et~al.(2022)Shu, Zhang, Gao, Yang, Wang and Gao}]{shu2022self}
\bibinfo{author}{Shu, X.}, \bibinfo{author}{Zhang, X.}, \bibinfo{author}{Gao,
  Q.}, \bibinfo{author}{Yang, M.}, \bibinfo{author}{Wang, R.},
  \bibinfo{author}{Gao, X.}, \bibinfo{year}{2022}.
\newblock \bibinfo{title}{Self-weighted anchor graph learning for multi-view
  clustering}.
\newblock \bibinfo{journal}{IEEE Transactions on Multimedia} .
\bibitem[{Tang et~al.(2023a)Tang, Li, Wang, Liu, Zhang and
  Zhu}]{DBLP:journals/tkde/TangLWLZZ23}
\bibinfo{author}{Tang, C.}, \bibinfo{author}{Li, Z.}, \bibinfo{author}{Wang,
  J.}, \bibinfo{author}{Liu, X.}, \bibinfo{author}{Zhang, W.},
  \bibinfo{author}{Zhu, E.}, \bibinfo{year}{2023}a.
\newblock \bibinfo{title}{Unified one-step multi-view spectral clustering}.
\newblock \bibinfo{journal}{{IEEE} Trans. Knowl. Data Eng.}
  \bibinfo{volume}{35}, \bibinfo{pages}{6449--6460}.
\bibitem[{Tang et~al.(2023b)Tang, Sun, Tang, Zheng, Liu, Huang and
  Zhang}]{DBLP:journals/nn/TangSTZLHZ23}
\bibinfo{author}{Tang, C.}, \bibinfo{author}{Sun, K.}, \bibinfo{author}{Tang,
  C.}, \bibinfo{author}{Zheng, X.}, \bibinfo{author}{Liu, X.},
  \bibinfo{author}{Huang, J.}, \bibinfo{author}{Zhang, W.},
  \bibinfo{year}{2023}b.
\newblock \bibinfo{title}{Multi-view subspace clustering via adaptive graph
  learning and late fusion alignment}.
\newblock \bibinfo{journal}{Neural Networks} \bibinfo{volume}{165},
  \bibinfo{pages}{333--343}.
\bibitem[{Tang et~al.(2022)Tang, Zheng, Liu, Zhang, Zhang, Xiong and
  Wang}]{DBLP:journals/tkde/TangZLZZXW22}
\bibinfo{author}{Tang, C.}, \bibinfo{author}{Zheng, X.}, \bibinfo{author}{Liu,
  X.}, \bibinfo{author}{Zhang, W.}, \bibinfo{author}{Zhang, J.},
  \bibinfo{author}{Xiong, J.}, \bibinfo{author}{Wang, L.},
  \bibinfo{year}{2022}.
\newblock \bibinfo{title}{Cross-view locality preserved diversity and consensus
  learning for multi-view unsupervised feature selection}.
\newblock \bibinfo{journal}{{IEEE} Trans. Knowl. Data Eng.}
  \bibinfo{volume}{34}, \bibinfo{pages}{4705--4716}.
\bibitem[{Tang et~al.(2023c)Tang, Zheng, Zhang, Liu, Zhu and
  Zhu}]{DBLP:journals/chinaf/TangZZLZZ23}
\bibinfo{author}{Tang, C.}, \bibinfo{author}{Zheng, X.},
  \bibinfo{author}{Zhang, W.}, \bibinfo{author}{Liu, X.}, \bibinfo{author}{Zhu,
  X.}, \bibinfo{author}{Zhu, E.}, \bibinfo{year}{2023}c.
\newblock \bibinfo{title}{Unsupervised feature selection via multiple graph
  fusion and feature weight learning}.
\newblock \bibinfo{journal}{Sci. China Inf. Sci.} \bibinfo{volume}{66}.
\bibitem[{Wang et~al.(2020)Wang, Xiao, Li, Wang, Chen and
  Fang}]{wang2020robust}
\bibinfo{author}{Wang, B.}, \bibinfo{author}{Xiao, Y.}, \bibinfo{author}{Li,
  Z.}, \bibinfo{author}{Wang, X.}, \bibinfo{author}{Chen, X.},
  \bibinfo{author}{Fang, D.}, \bibinfo{year}{2020}.
\newblock \bibinfo{title}{Robust self-weighted multi-view projection
  clustering}, in: \bibinfo{booktitle}{Proceedings of the AAAI Conference on
  Artificial Intelligence}, pp. \bibinfo{pages}{6110--6117}.
\bibitem[{Wang et~al.(2016)Wang, Fu, Hao, Tao and Wu}]{wang2016scalable}
\bibinfo{author}{Wang, M.}, \bibinfo{author}{Fu, W.}, \bibinfo{author}{Hao,
  S.}, \bibinfo{author}{Tao, D.}, \bibinfo{author}{Wu, X.},
  \bibinfo{year}{2016}.
\newblock \bibinfo{title}{Scalable semi-supervised learning by efficient anchor
  graph regularization}.
\newblock \bibinfo{journal}{IEEE Transactions on Knowledge and Data
  Engineering} \bibinfo{volume}{28}, \bibinfo{pages}{1864--1877}.
\bibitem[{Wang et~al.(2021)Wang, Liu, Zhu, Zhang, Zhang, Gao and
  Zhu}]{wang2021fast}
\bibinfo{author}{Wang, S.}, \bibinfo{author}{Liu, X.}, \bibinfo{author}{Zhu,
  X.}, \bibinfo{author}{Zhang, P.}, \bibinfo{author}{Zhang, Y.},
  \bibinfo{author}{Gao, F.}, \bibinfo{author}{Zhu, E.}, \bibinfo{year}{2021}.
\newblock \bibinfo{title}{Fast parameter-free multi-view subspace clustering
  with consensus anchor guidance}.
\newblock \bibinfo{journal}{IEEE Transactions on Image Processing}
  \bibinfo{volume}{31}, \bibinfo{pages}{556--568}.
\bibitem[{Wang et~al.(2017)Wang, Guo, Lei, Zhang and
  Li}]{DBLP:conf/cvpr/WangGLZL17}
\bibinfo{author}{Wang, X.}, \bibinfo{author}{Guo, X.}, \bibinfo{author}{Lei,
  Z.}, \bibinfo{author}{Zhang, C.}, \bibinfo{author}{Li, S.Z.},
  \bibinfo{year}{2017}.
\newblock \bibinfo{title}{Exclusivity-consistency regularized multi-view
  subspace clustering}, in: \bibinfo{booktitle}{2017 {IEEE} Conference on
  Computer Vision and Pattern Recognition, {CVPR} 2017},
  \bibinfo{publisher}{{IEEE} Computer Society}. pp. \bibinfo{pages}{1--9}.
\bibitem[{Wang et~al.(2015)Wang, Lin, Wu, Zhang, Zhang and
  Huang}]{DBLP:journals/tip/WangLWZZH15}
\bibinfo{author}{Wang, Y.}, \bibinfo{author}{Lin, X.}, \bibinfo{author}{Wu,
  L.}, \bibinfo{author}{Zhang, W.}, \bibinfo{author}{Zhang, Q.},
  \bibinfo{author}{Huang, X.}, \bibinfo{year}{2015}.
\newblock \bibinfo{title}{Robust subspace clustering for multi-view data by
  exploiting correlation consensus}.
\newblock \bibinfo{journal}{{IEEE} Trans. Image Process.} \bibinfo{volume}{24},
  \bibinfo{pages}{3939--3949}.
\bibitem[{Wang et~al.(2018)Wang, Wu, Lin and Gao}]{DBLP:journals/tnn/WangWLG18}
\bibinfo{author}{Wang, Y.}, \bibinfo{author}{Wu, L.}, \bibinfo{author}{Lin,
  X.}, \bibinfo{author}{Gao, J.}, \bibinfo{year}{2018}.
\newblock \bibinfo{title}{Multiview spectral clustering via structured low-rank
  matrix factorization}.
\newblock \bibinfo{journal}{{IEEE} Trans. Neural Networks Learn. Syst.}
  \bibinfo{volume}{29}, \bibinfo{pages}{4833--4843}.
\bibitem[{Wei et~al.(2023)Wei, Yue, Feng, Cui and
  Liang}]{DBLP:journals/tkde/WeiYFCL23}
\bibinfo{author}{Wei, W.}, \bibinfo{author}{Yue, Q.}, \bibinfo{author}{Feng,
  K.}, \bibinfo{author}{Cui, J.}, \bibinfo{author}{Liang, J.},
  \bibinfo{year}{2023}.
\newblock \bibinfo{title}{Unsupervised dimensionality reduction based on fusing
  multiple clustering results}.
\newblock \bibinfo{journal}{{IEEE} Trans. Knowl. Data Eng.}
  \bibinfo{volume}{35}, \bibinfo{pages}{3211--3223}.
\bibitem[{Winn and Jojic(2005)}]{winn2005locus}
\bibinfo{author}{Winn, J.}, \bibinfo{author}{Jojic, N.}, \bibinfo{year}{2005}.
\newblock \bibinfo{title}{Locus: Learning object classes with unsupervised
  segmentation}, in: \bibinfo{booktitle}{Tenth IEEE International Conference on
  Computer Vision (ICCV'05) Volume 1}, \bibinfo{organization}{IEEE}. pp.
  \bibinfo{pages}{756--763}.
\bibitem[{Xia et~al.(2014)Xia, Pan, Du and Yin}]{DBLP:conf/aaai/XiaPDY14}
\bibinfo{author}{Xia, R.}, \bibinfo{author}{Pan, Y.}, \bibinfo{author}{Du, L.},
  \bibinfo{author}{Yin, J.}, \bibinfo{year}{2014}.
\newblock \bibinfo{title}{Robust multi-view spectral clustering via low-rank
  and sparse decomposition}, in: \bibinfo{booktitle}{Proceedings of the
  Twenty-Eighth {AAAI} Conference on Artificial Intelligence, July 27 -31,
  2014}, \bibinfo{publisher}{{AAAI} Press}. pp. \bibinfo{pages}{2149--2155}.
\bibitem[{Xu et~al.(2013)Xu, Tao and Xu}]{DBLP:journals/corr/abs-1304-5634}
\bibinfo{author}{Xu, C.}, \bibinfo{author}{Tao, D.}, \bibinfo{author}{Xu, C.},
  \bibinfo{year}{2013}.
\newblock \bibinfo{title}{A survey on multi-view learning}.
\newblock \bibinfo{journal}{CoRR} \bibinfo{volume}{abs/1304.5634}.
\bibitem[{Yan et~al.(2023a)Yan, Gu, Ren, Yue, Liu, Xu and
  Lin}]{DBLP:journals/inffus/YanGR0LXL23}
\bibinfo{author}{Yan, W.}, \bibinfo{author}{Gu, M.}, \bibinfo{author}{Ren, J.},
  \bibinfo{author}{Yue, G.}, \bibinfo{author}{Liu, Z.}, \bibinfo{author}{Xu,
  J.}, \bibinfo{author}{Lin, W.}, \bibinfo{year}{2023}a.
\newblock \bibinfo{title}{Collaborative structure and feature learning for
  multi-view clustering}.
\newblock \bibinfo{journal}{Inf. Fusion} \bibinfo{volume}{98},
  \bibinfo{pages}{101832}.
\bibitem[{Yan et~al.(2023b)Yan, Gu, Ren, Yue, Liu, Xu and
  Lin}]{yan2023collaborative}
\bibinfo{author}{Yan, W.}, \bibinfo{author}{Gu, M.}, \bibinfo{author}{Ren, J.},
  \bibinfo{author}{Yue, G.}, \bibinfo{author}{Liu, Z.}, \bibinfo{author}{Xu,
  J.}, \bibinfo{author}{Lin, W.}, \bibinfo{year}{2023}b.
\newblock \bibinfo{title}{Collaborative structure and feature learning for
  multi-view clustering}.
\newblock \bibinfo{journal}{Information Fusion} \bibinfo{volume}{98},
  \bibinfo{pages}{101832}.
\bibitem[{Yan et~al.(2022)Yan, Xu, Liu, Yue and Tang}]{DBLP:conf/mm/YanXLYT22}
\bibinfo{author}{Yan, W.}, \bibinfo{author}{Xu, J.}, \bibinfo{author}{Liu, J.},
  \bibinfo{author}{Yue, G.}, \bibinfo{author}{Tang, C.}, \bibinfo{year}{2022}.
\newblock \bibinfo{title}{Bipartite graph-based discriminative feature learning
  for multi-view clustering}, in: \bibinfo{booktitle}{{MM} '22: The 30th {ACM}
  International Conference on Multimedia, 2022}, \bibinfo{publisher}{{ACM}}.
  pp. \bibinfo{pages}{3403--3411}.
\bibitem[{Yan et~al.(2023c)Yan, Zhang, Lv, Tang, Yue, Liao and
  Lin}]{DBLP:conf/cvpr/YanZLT0LL23}
\bibinfo{author}{Yan, W.}, \bibinfo{author}{Zhang, Y.}, \bibinfo{author}{Lv,
  C.}, \bibinfo{author}{Tang, C.}, \bibinfo{author}{Yue, G.},
  \bibinfo{author}{Liao, L.}, \bibinfo{author}{Lin, W.}, \bibinfo{year}{2023}c.
\newblock \bibinfo{title}{Gcfagg: Global and cross-view feature aggregation for
  multi-view clustering}, in: \bibinfo{booktitle}{{IEEE/CVF} Conference on
  Computer Vision and Pattern Recognition, {CVPR}, 2023},
  \bibinfo{publisher}{{IEEE}}. pp. \bibinfo{pages}{19863--19872}.
\bibitem[{Yang et~al.(2023)Yang, Wu, Zhang, Lin, Nie and Chen}]{yang2023robust}
\bibinfo{author}{Yang, B.}, \bibinfo{author}{Wu, J.}, \bibinfo{author}{Zhang,
  X.}, \bibinfo{author}{Lin, Z.}, \bibinfo{author}{Nie, F.},
  \bibinfo{author}{Chen, B.}, \bibinfo{year}{2023}.
\newblock \bibinfo{title}{Robust anchor-based multi-view clustering via
  spectral embedded concept factorization}.
\newblock \bibinfo{journal}{Neurocomputing} .
\bibitem[{Yang et~al.(2022a)Yang, Zhang, Li, Nie and Wang}]{yang2022efficient}
\bibinfo{author}{Yang, B.}, \bibinfo{author}{Zhang, X.}, \bibinfo{author}{Li,
  Z.}, \bibinfo{author}{Nie, F.}, \bibinfo{author}{Wang, F.},
  \bibinfo{year}{2022}a.
\newblock \bibinfo{title}{Efficient multi-view k-means clustering with multiple
  anchor graphs}.
\newblock \bibinfo{journal}{IEEE Transactions on Knowledge and Data
  Engineering} .
\bibitem[{Yang et~al.(2022b)Yang, Zhang, Nie and Wang}]{yang2022fast}
\bibinfo{author}{Yang, B.}, \bibinfo{author}{Zhang, X.}, \bibinfo{author}{Nie,
  F.}, \bibinfo{author}{Wang, F.}, \bibinfo{year}{2022}b.
\newblock \bibinfo{title}{Fast multiview clustering with spectral embedding}.
\newblock \bibinfo{journal}{IEEE Transactions on Image Processing}
  \bibinfo{volume}{31}, \bibinfo{pages}{3884--3895}.
\bibitem[{Yang et~al.(2022c)Yang, Gao, Xia, Yang and Gao}]{yang2022multiview}
\bibinfo{author}{Yang, H.}, \bibinfo{author}{Gao, Q.}, \bibinfo{author}{Xia,
  W.}, \bibinfo{author}{Yang, M.}, \bibinfo{author}{Gao, X.},
  \bibinfo{year}{2022}c.
\newblock \bibinfo{title}{Multiview spectral clustering with bipartite graph}.
\newblock \bibinfo{journal}{IEEE Transactions on Image Processing}
  \bibinfo{volume}{31}, \bibinfo{pages}{3591--3605}.
\bibitem[{Zhang et~al.(2018)Zhang, Liu, Shen, Shen and Shao}]{zhang2018binary}
\bibinfo{author}{Zhang, Z.}, \bibinfo{author}{Liu, L.}, \bibinfo{author}{Shen,
  F.}, \bibinfo{author}{Shen, H.T.}, \bibinfo{author}{Shao, L.},
  \bibinfo{year}{2018}.
\newblock \bibinfo{title}{Binary multi-view clustering}.
\newblock \bibinfo{journal}{IEEE transactions on pattern analysis and machine
  intelligence} \bibinfo{volume}{41}, \bibinfo{pages}{1774--1782}.
\bibitem[{Zhao et~al.(2023)Zhao, Yang and Nie}]{zhao2023auto}
\bibinfo{author}{Zhao, M.}, \bibinfo{author}{Yang, W.}, \bibinfo{author}{Nie,
  F.}, \bibinfo{year}{2023}.
\newblock \bibinfo{title}{Auto-weighted orthogonal and nonnegative graph
  reconstruction for multi-view clustering}.
\newblock \bibinfo{journal}{Information Sciences} \bibinfo{volume}{632},
  \bibinfo{pages}{324--339}.
\bibitem[{Zhou et~al.(2021)Zhou, Ou, Liu, Wang, Liu, Wang, Zhu, Yin and
  Xu}]{zhou2021multiple}
\bibinfo{author}{Zhou, S.}, \bibinfo{author}{Ou, Q.}, \bibinfo{author}{Liu,
  X.}, \bibinfo{author}{Wang, S.}, \bibinfo{author}{Liu, L.},
  \bibinfo{author}{Wang, S.}, \bibinfo{author}{Zhu, E.}, \bibinfo{author}{Yin,
  J.}, \bibinfo{author}{Xu, X.}, \bibinfo{year}{2021}.
\newblock \bibinfo{title}{Multiple kernel clustering with compressed subspace
  alignment}.
\newblock \bibinfo{journal}{IEEE Transactions on Neural Networks and Learning
  Systems} .

\end{thebibliography}
	

		
		
		
\end{document}